%% file: main.tex
\documentclass[sigplan,screen]{acmart}

\copyrightyear{2025}
\acmYear{2025}
\setcopyright{cc}
\setcctype{by}
\acmConference[ASPLOS '25] {Proceedings of the 30th ACM International Conference on Architectural Support for Programming Languages and Operating Systems, Volume 2}{March 30--April 3, 2025}{Rotterdam, Netherlands.}
\acmBooktitle{Proceedings of the 30th ACM International Conference on Architectural Support for Programming Languages and Operating Systems, Volume 2 (ASPLOS '25), March 30--April 3, 2025, Rotterdam, Netherlands}
\acmISBN{979-8-4007-1079-7/25/03}
\acmDOI{10.1145/3676641.3716249}

\usepackage[]{hyperref}

\usepackage{float}
\usepackage{xspace}
\usepackage{soul}
\usepackage{tabularray}
\usepackage{xcolor}
\usepackage{multirow}
\usepackage{multicol}
\usepackage{makecell}
\usepackage{algorithm}
\usepackage{algorithmic}
\usepackage{balance}
\UseTblrLibrary{booktabs}
\DefTblrTemplate{caption-tag}{default}{\textbf{Table\hspace{0.25em}\thetable}}
\DefTblrTemplate{caption-sep}{default}{\textbf{.}\enskip}
\DefTblrTemplate{caption-text}{default}{\InsertTblrText{caption}}

\begin{document}

\newcommand{\sys}{Relax\xspace}
\newcommand{\papertitle}{Composable Abstractions for End-to-End Dynamic Machine Learning}
\title{\sys: \papertitle}

\author{Ruihang Lai}
\authornotemark[1]
\affiliation{%
  \institution{Carnegie Mellon University}
}

\author{Junru Shao}
\authornotemark[1] 
\authornotemark[2]
\affiliation{%
  \institution{OpenAI}  
}

\author{Siyuan Feng}
\authornotemark[1]
\affiliation{%
  \institution{Shanghai Jiao Tong University}
}

\author{Steven Lyubomirsky}
\authornote{Equal contribution.} 
\authornotemark[2]
\affiliation{%
  \institution{NVIDIA}
}

\author{Bohan Hou}
\affiliation{%
  \institution{Carnegie Mellon University}
}

\author{Wuwei Lin}
\authornotemark[2]
\affiliation{%
  \institution{OpenAI}  
}

\author{Zihao Ye}
\affiliation{%
  \institution{University of Washington}
}

\author{Hongyi Jin}
\affiliation{%
  \institution{Carnegie Mellon University}
}

\author{Yuchen Jin}
\authornotemark[2]
\affiliation{%
  \institution{Hyperbolic}
}

\author{Jiawei Liu}
\authornotemark[2]
\affiliation{%
  \institution{University of Illinois Urbana-Champaign}
}

\author{Lesheng Jin}
\authornotemark[2]
\affiliation{%
  \institution{Hyperbolic}
}

\author{Yaxing Cai}
\authornotemark[2]
\affiliation{%
  \institution{NVIDIA}
}

\author{Ziheng Jiang}
\authornotemark[2]
\affiliation{%
  \institution{ByteDance}
}

\author{Yong Wu}
\authornotemark[2]
\affiliation{%
  \institution{NVIDIA}
}

\author{Sunghyun Park}
\authornotemark[2]
\affiliation{%
  \institution{NVIDIA}
}

\author{Prakalp Srivastava}
\authornotemark[2]
\affiliation{%
  \institution{Netflix}
}

\author{Jared Roesch}
\authornote{Work done while at OctoAI (now acquired by NVIDIA).}
\affiliation{%
  \institution{NVIDIA}
}

\author{Todd C. Mowry}
\affiliation{%
  \institution{Carnegie Mellon University}
}

\author{Tianqi Chen}
\affiliation{%
  \institution{Carnegie Mellon University}
}
\affiliation{%
  \institution{NVIDIA}
}

\renewcommand{\shortauthors}{Ruihang Lai et al.}

\ccsdesc[500]{Software and its engineering~Domain specific languages}

\keywords{Machine Learning Compiler; Dynamic-Shape Machine Learning}

\input{sections/macros}
\input{sections/abstract}

\maketitle 

\input{sections/introduction}

\input{sections/overview}
\input{sections/abstraction}

\input{sections/optimizations}

\input{sections/evaluation}

\input{sections/related-work}
\input{sections/conclusion}
\input{sections/acknowledgement}

\bibliographystyle{plain}
\balance
\bibliography{references}

\end{document}

%% file: sections/macros.tex
\newcommand{\matchcast}{\texttt{match\_cast}}
\newcommand{\calltir}{\texttt{call\_tir}}
\newcommand{\calllibrary}{\texttt{call\_dps\_library}}

\newcommand{\todo}[1]{{\color{red}TODO: {#1}}}
\newcommand{\hint}[1]{{\color{blue}Hint: {#1}}}
\newcommand{\keypoint}[1]{{\textbf{\color{red}Key point: {#1}.}}}
\newcommand{\keypointcomment}[1]{}

\newcommand{\vspacebeforecap}{\vspace{-2em}}
\newcommand{\vspaceaftercap}{\vspace{-1em}}

\newcommand{\MyPara}[1]{\vspace{.0em}\noindent\textbf{#1}}

\newcommand{\squishlist}{
   \begin{list}{$\bullet$}
    { \setlength{\itemsep}{1pt}      \setlength{\parsep}{3pt}
      \setlength{\topsep}{3pt}       \setlength{\partopsep}{0pt}
      \setlength{\leftmargin}{1em} \setlength{\labelwidth}{1em}
      \setlength{\labelsep}{0.5em} } }

\newcommand{\squishlisttwo}{
   \begin{list}{$\bullet$}
    { \setlength{\itemsep}{0pt}    \setlength{\parsep}{0pt}
      \setlength{\topsep}{0pt}     \setlength{\partopsep}{0pt}
      \setlength{\leftmargin}{2em} \setlength{\labelwidth}{1.5em}
      \setlength{\labelsep}{0.5em} } }

\newcommand{\squishend}{
    \end{list}  }

%% file: sections/abstract.tex
\begin{abstract}

Dynamic shape computations have become critical in modern machine learning workloads, especially in emerging large language models. The success of these models has driven the demand for their universal deployment across a diverse set of backend environments. In this paper, we present \sys{}, a compiler abstraction for optimizing end-to-end dynamic machine learning workloads. 
\sys{} introduces a cross-level abstraction that encapsulates computational graphs, loop-level tensor programs and external libraries calls in a single representation.
\sys{} also introduces first-class symbolic shape annotations to track dynamic shape computations globally across the
program, enabling dynamic shape--aware cross-level optimizations.
We build an end-to-end compilation framework using the proposed approach to optimize dynamic shape models. Experimental results on LLMs show that \sys{} delivers performance competitive  with  state-of-the-art systems across various GPUs and enables deployment of emerging models to a broader set of emerging environments, including mobile phones, embedded devices, and web browsers.


\end{abstract}

%% file: sections/introduction.tex
\begin{figure*}[t]
    \centering
    \includegraphics[width=.84\textwidth]{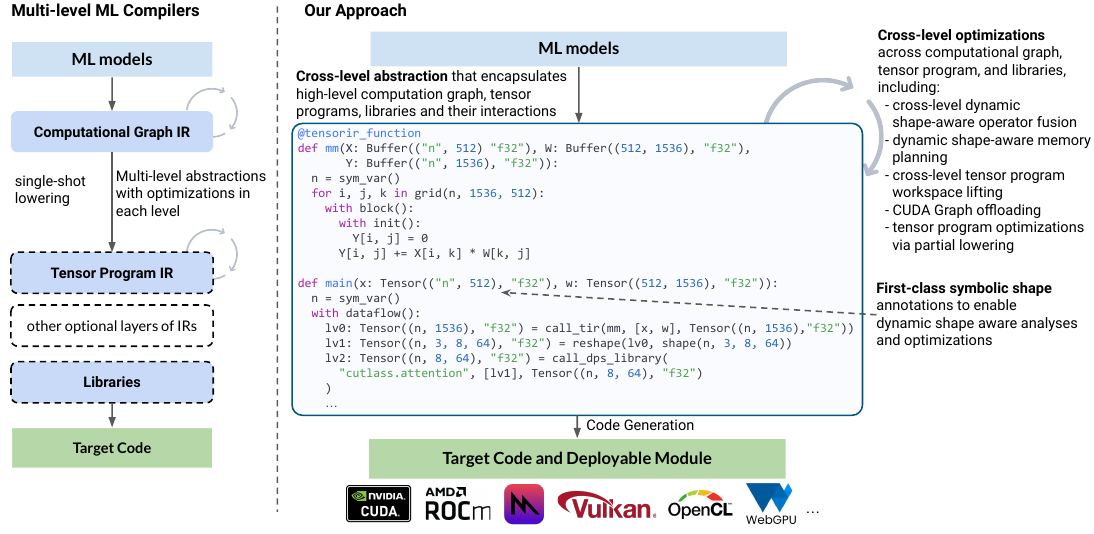}
    \caption{Overview of our approach.
    We also present a cross-level abstraction that encapsulates
    the computational graph,
    foreign tensor program and the external library function levels.
    We introduce first-class symbolic shape annotations
    to track dynamic shape computations globally across the program,
    and enable
    dynamic shape--aware optimizations across levels.
    }
    \vspaceaftercap
    \label{fig:overview}
\end{figure*}

\section{Introduction}
\label{sec:introduction}

Machine learning (ML) applications are now ubiquitous in everyday life and the broader economy.
The arrival of GPT-4 and open-source large language models~(LLMs)~\cite{touvron2023llama,rozière2024code,black2022gptneox20b,NIPS2017_3f5ee243,xu2023wizardlm}
has created promising opportunities for building even more powerful modern AI systems for processing text, images, audio, and more. The success of these models has also led to growing demand for their deployment across a diverse set of backend environments, including
servers, personal computers, vehicles, and mobile devices.

Machine learning frameworks~\cite{abadi2016tensorflow,paszke2019pytorch} are responsible for deploying many of these models to diverse backends.
ML compilers~\cite{chen2018tvm,lattner2021mlir,iree,pytorch2} aim to reduce the gap between model
and backends by ingesting model computations into common program abstractions, performing optimizations such as 
operator fusion, and generating high-performance code on a diverse set of platforms.
However, much engineering effort is required to support the tensor operators in a model for different hardware backends,
particularly since
most of these models make use of dynamic tensor shapes. Dynamic shapes contain variables that may depend on program values, which makes it more difficult to perform crucial optimizations like static memory planning.
For example, language models must handle variable-sized input messages, KV-cache context length, vocabulary sizes, and other sources of shape dynamism.

Typical ML compilers include multiple levels of abstractions (or intermediate representations) and single shot lowerings 
across levels.
The high-level computational graph ~\cite{Roesch_2018,torchfx,lattner2021mlir} describes a model
using dataflow graph of high-level tensor operators (e.g., \texttt{matmul}, \texttt{reshape}) to facilitate global
computation rewrites.
Tensor programs~\cite{hagedorn2023graphene,tillet2019triton,feng2023tensorir} describe tensor operators
with low-level loops and indexed buffer accesses,
to enable fine-grained kernel optimizations (such as loop tiling) at this level.
Operator libraries~\cite{chetlur2014cudnn,Thakkar_CUTLASS_2023}
allow for offloading a tensor operator to vendor optimized routines.

Most end-to-end ML compilers use computational graphs as the high-level representation and treat tensor programs and operator libraries as \emph{foreign functions} that are usually opaque.
Dynamic shapes are usually handled at each level within each function. Relay~\cite{Roesch_2018} and IREE~\cite{lattner2021mlir} handles dynamic shapes in computational graphs through ``unknown'' annotations and do not track relations between dynamic shapes. The PyTorch compiler~\cite{pytorch2} enables just-in-time~(JIT) graph tracing and handles dynamic shape tracking within each traced function. The JIT approach eliminates the need for shape tracking across function boundaries, but also limits its applications on emerging platforms with constrained environments, such as mobile and WebGPU. DietCode~\cite{MLSYS2022_f89b79c9}, CoRA~\cite{MLSYS2022_afe8a457}, and SparseTIR~\cite{ye2023sparsetir}
focus on optimizing dynamic shapes within each tensor program. 
Halide~\cite{ragan2013halide}  tracks dynamic shapes in tensor programs and provides primitives to call external library functions from tensor programs.

Despite the developments in dynamic shape handling at each level and function, challenges still exist in optimizations across these levels and functions. First, as we target a broad set of emerging platforms, we must enable ahead-of-time~(AOT) compilation, which necessitates full program optimizations across functions. 
Additionally, user-defined operators such as customized quantization decode require
computational graph optimizations to be aware of foreign functions.
Finally, single-shot lowering makes it harder to analyze or transform tensor programs first and then use the results to inform high-level graph optimizations. 
Dynamic shape tracking threads through all these challenges in every incremental transformation, as losing the shape relation information can significantly hinder the optimizations we can perform across operators and functions in the program.

To address these challenges, we introduce \sys{}, a holistic
AOT
compiler program abstraction for emerging end-to-end dynamic machine learning models on emerging platforms.
\sys{} enables an abstraction
that encapsulates levels of computational graphs, loop-level tensor programs and external library functions 
at the same time,
which we call \emph{cross-level abstraction} 
in this paper.\footnote{We leverage other lower layers in ML compilers for
GPU source code generation.}
We also introduce \emph{first-class symbolic shapes} to track and represent the relations of dynamic
shape dimensions.
\sys{} uses variables to represent symbolic shape dimensions and employs symbolic deductions to track dynamic shapes across 
tensor operators, subgraph function calls
and foreign function calls of tensor programs and external library functions
statically when possible, with a dynamic fallback as needed.
The cross-level abstraction with first-class symbolic shape
allows for analyses and optimizations across
these abstraction levels,
and meanwhile preserves symbolic shape information in IR during optimizations.
We introduce a collection of
optimizations to enable dynamic shape-aware operator fusion,
tensor program workspace lifting,
memory optimization,  graph offloading and tensor operator optimizations.
Finally, we build an end-to-end compilation framework on top of these
elements. This paper makes the following contributions:

\squishlist
    \item We propose a design for cross-level abstractions
    to enable optimizations and analyses across the traditional levels of abstractions in ML frameworks.
    \item We present a program abstraction with a first-class symbolic shape approach that tracks dynamic shape relations globally across tensor operators, subgraph function calls and foreign function calls of tensor programs and external libraries,
    enabling full-program symbolic shape tracking and
    dynamic shape--aware optimizations across abstraction levels.
    \item We build an AOT end-to-end compilation framework to enable full program deployment of emerging models to diverse hardware backends, including many emerging backends that are not well supported by established frameworks.
\squishend

Experimental results show that \sys{} compiles and optimizes emerging LLMs onto a broad set of emerging devices and environments, including mobile phones, embedded devices, and web browsers via WebGPU. 
Additionally, \sys{} delivers competitive performance to heavily optimized platform-specific solutions. \sys{} is incorporated into a major open-source project and enables support for the universal deployment of emerging machine learning models.

%% file: sections/overview.tex
\vspace{-3pt}
\section{Overview}
\label{sec:overview}

This section describes the key insights of our approach and gives an overview of the paper. 
\autoref{fig:overview} summarizes our overall approach, focusing on 
two key
designs that enable compiler optimizations across all levels of abstractions for dynamic machine learning models.

First, we observe that ML compilers often need to
go through several abstraction levels to bring
a machine learning model to a target platform. 
Typical layers include
computational graphs,
tensor programs and external libraries.
Traditionally, ML compilers focus on optimizations within each individual
abstraction level and do a uni-directional single-shot lowering
from one level to the next level.
\sys{} brings computational graphs,
tensor programs, and libraries into a single unified cross-level abstraction,
allowing
the interaction of those components
with foreign function calls
to tensor programs and external library functions.
This design allows us to
incrementally optimize or partially lower portions of the computation
using different approaches, with analyses from all abstraction levels
taken into account.

Second, we observe that while emerging machine learning workloads involve dynamic shape computations, 
we can perform a substantial amount of static analyses and optimizations by considering the relations between shapes. Thus, we introduce annotations that can track the shapes of intermediate computations through symbolic variables and symbolic shape computations. 
Our approach globally tracks these dynamic shape computations across
subgraph function calls 
and foreign function calls of the tensor program and external library levels
to represent dynamic shapes throughout 
the program
and enable dynamic shape--aware optimizations.

We introduce \sys{}'s abstraction design in \S\ref{sec:abstraction},
and discuss a concrete set of cross-level compiler optimizations enabled by
our design in \S\ref{sec:composable-optimizations}.

%% file: sections/abstraction.tex
\section{\sys{} Abstraction}
\label{sec:abstraction}

This section introduces the overall \sys{} abstraction.
We start with the language constructs, followed by
the first-class symbolic shape and cross-level abstractions
in \sys{}.

\subsection{Language Constructs}
\label{sec:language-construct}

\begin{figure}[t]
    \centering
    \includegraphics[width=0.48\textwidth]{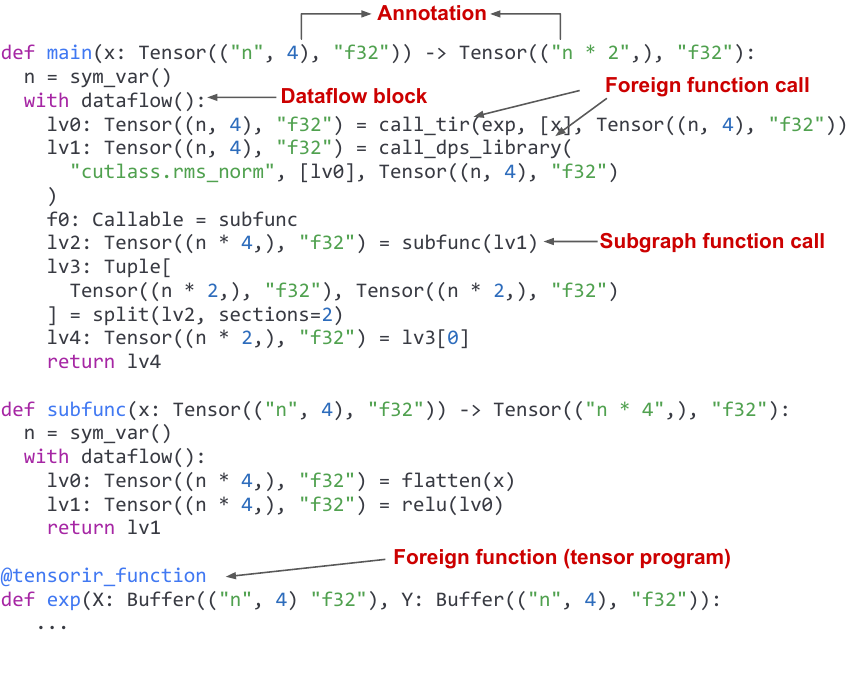}
    \vspacebeforecap
    \vspace{-1.0em}
    \caption{Key elements of \sys{} abstraction.}
    \vspaceaftercap
    \label{fig:approach:language-constructs}
\end{figure}

\begin{table}[t]
    \centering
    \caption{Annotations in \sys{} with examples.}
    \vspace{-0.5em}
    \includegraphics[width=0.48\textwidth]{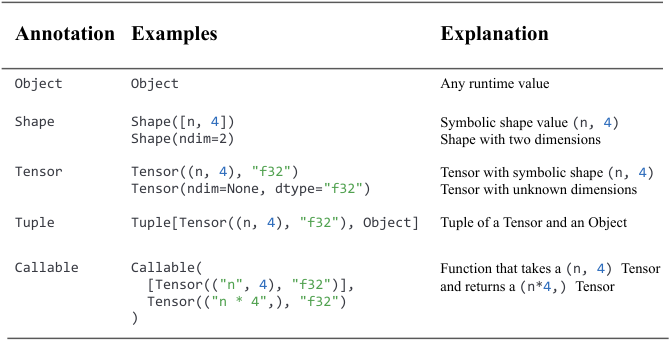}
    \vspaceaftercap
    \vspace{-1.5em}
    \label{table:relax-annotations}
\end{table}

\sys{} is an imperative compiler abstraction with first-class functions, focusing on operating tensors at a high level (the ``graph level,'' as most ML compilers refer to it). \sys{} programs apply high-level operators to entire tensors and can pass tensors (or tuples of tensors) among functions, or invoke lower-level tensor programs
or external library functions
for loop-level operations on tensors.
This section describes three main elements of \sys{}: structural annotations, dataflow blocks, and function calls both within and across levels of abstractions.
These constructs bring the whole system together
by offering symbolic shape guidance in compiler optimizations of joint transformations across
graph level and foreign tensor program level,
and meanwhile preserving the symbolic shape information in the program during these transformations.

\MyPara{Annotations.}
Each value in \sys{} is associated with an annotation that conveys structural information, similar to a static type. \autoref{table:relax-annotations} summarizes different annotations in \sys{} and their usage examples and explanations respectively.
We design the annotation syntax to be embeddable in Python AST, and quote symbolic expressions into strings (e.g., \texttt{"n*4"}) in function signatures, when the symbolic variables are yet to be declared.\footnote{The quoted strings can be read as normal symbolic expressions. This syntax simplifies parsing in python ast and can be changed.}
To enrich the shape expressiveness of annotations and ensure the shape annotation coverage,
we reuse the loop-level tensor program expression system, so that shape annotations support all integer expressions that tensor programs support, and the compiler symbolic expression analyses
(e.g., expression equality proof) can take advantage of
common expressions.
Annotations indicate at compile time the overall types (e.g., tensor, tuple) and additional information about values, such as the shape and data type of a tensor. Annotations form the basis for first-class symbolic shapes and cross-level abstractions.

\MyPara{Dataflow Blocks.}
A dataflow block in \sys{} demarcates a side effect--free program (sub-)region without control flows, i.e., a straight-line sequence of pure operations, in order to simplify program transformations.  For example, when performing dead code elimination over \sys{} dataflow blocks, one can safely remove unused operators without having to consider whether this could affect the visible behavior of the program by removing an operation with side effects. 

\MyPara{Function Calls.}
\sys{} incorporates function calls that can be within the same level of abstraction (i.e., allowing one graph-level function to invoke another graph-level function) or across levels of abstraction, namely allowing graph-level functions to call 
foreign
tensor program functions and external libraries.
Calling 
foreign
loop-level tensor programs and external library functions serves as a foundational element for cross-level abstractions, as explored in detail in \S\ref{sec:cross-level-abstraction}.
We use TensorIR~\cite{feng2023tensorir} as the loop-level tensor program abstraction, though the same principle can be used for other loop-level abstractions.

\subsection{First-Class Symbolic Shape Abstraction}
\label{sec:first-class-sym-shape}

\begin{figure}[t]
    \centering
    \includegraphics[width=0.48\textwidth]{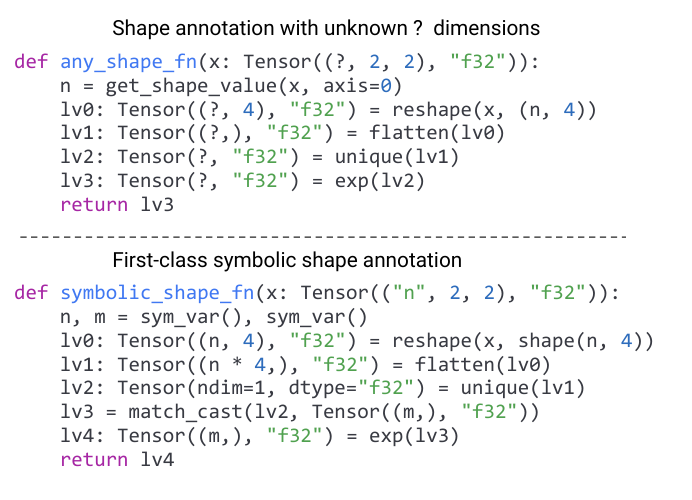}
    \vspacebeforecap
    \caption{Comparison of first-class symbolic shape annotation with unknown dynamic shape annotation. First-class symbolic shape  enables comprehensive symbolic analysis and facilitates advanced dynamic shape--aware optimizations.}
    \label{fig:approach:first-class-sym-shape}
    \vspaceaftercap
\end{figure}

The shape of a tensor is very useful information in the context of ML frameworks, especially for memory planning.
Oftentimes, however, tensor shapes are unknown at compile time and must be dealt with dynamically.
One approach for reasoning about dynamic
shape dimensions is to introduce an \emph{any}~(or unknown) value to represent dynamic
dimensions, as in ONNX~\cite{onnx}, Relay~\cite{roesch2020thesis}, and some MLIR dialects~\cite{lattner2021mlir}. 
Unfortunately, this approach fails to preserve potentially useful information, like relations or constraints between shape dimensions (e.g., if one tensor has dimension $n$, another may be $4n$). 
Such information is
valuable for compiler optimizations, whereas marking the dimension as \emph{any} erases it entirely.

We instead introduce a first-class symbolic shape annotation~(shown in \autoref{fig:approach:first-class-sym-shape}) for better reasoning and optimizations of dynamic shape models.
A symbolic shape annotation describes each shape dimension using a symbolic expression, comprised of integer variables and constants.
Consequently, for fully static models, \sys shape annotations subsumes existing static shape--based annotations.
For models with a mixed set of dynamic and static dimensions, \sys  not only expresses the shape dimensions symbolically but tracks the symbolic relations over dimensions as well. 
These symbolic shape relations help us apply more dynamic shape--aware optimizations.
For example, we will know the total number of elements after the \texttt{flatten} operator is $4n$ in \autoref{fig:approach:first-class-sym-shape} and is the same as the
input, suggesting potential buffer reuses.

Besides serving as annotations, a symbolic shape can also be used
as a first-class value in the computation. For example, the \texttt{reshape} operator
in \autoref{fig:approach:first-class-sym-shape}  takes the symbolic shape $(n, 4)$
as input and outputs a tensor value with the same shape.
Symbolic shape expressions can also be used to construct arguments to tensor functions.

It is not always possible to track the shape relations at compile time.
We cannot deduce the output shape of data-dependent operators, e.g., \texttt{unique} in \autoref{fig:approach:first-class-sym-shape}, whose output tensor shape depends on the runtime values of its input.
For such cases, we provide coarse-grained annotations~(e.g., \texttt{Shape(ndim=2)}
in \autoref{table:relax-annotations}, meaning that the shape has two dimensions but both are unknown). 
To reason about such cases, we introduce a special construct \matchcast{} that 
asserts a symbolic shape annotation for a value, allowing for the introduction of new symbolic variables.
The compiler inserts runtime checks for each \matchcast{}, throwing an error if a constraint is violated.
In our particular example in \autoref{fig:approach:first-class-sym-shape},
though the compiler does not know
the shape of \texttt{lv2}
after the \texttt{unique} operator,
one can use \matchcast{} to assume that it has shape
$(m,)$ (as in its alias \texttt{lv3}).
\matchcast{} can be inserted by both front-ends and compiler 
passes to suggest more specific symbolic shapes and
serves as a valuable tool for developers to indicate shape information within programs.

\begin{figure}[t]
    \centering
    \includegraphics[width=0.48\textwidth]{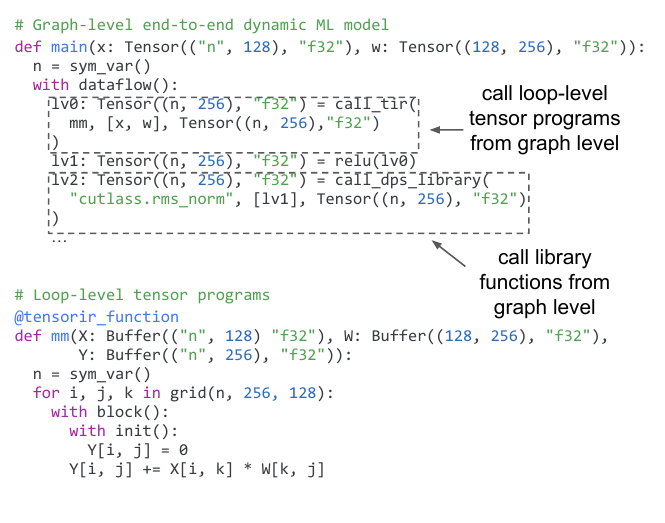}
    \vspacebeforecap
    \vspace{-1em}
    \caption{Cross-level abstractions: Graph-level function calls and communicates with loop-level TensorIR using \calltir{}, and invokes library functions via \calllibrary{}.} 
    \vspaceaftercap
    \label{fig:approach:cross-level-example}
\end{figure}

\begin{figure}[!t]
    \centering
    \vspace{0.5em}
    \includegraphics[width=0.45\textwidth]{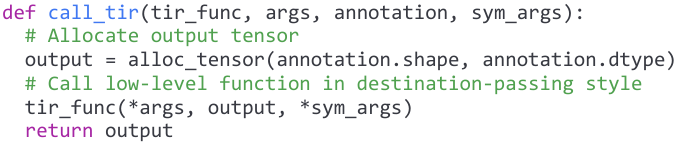}
    \caption{The semantics explanation of \calltir{}.} 
    \vspaceaftercap
    \label{fig:approach:calltir-semantics}
\end{figure}
\subsection{Cross-Level Abstraction}
\label{sec:cross-level-abstraction}

This section describes the constructs in \sys{} that 
enable cross-level abstractions. Our main goal is to design primitives that naturally represent and optimize interactions
of computational graphs, foreign tensor programs and libraries.

To achieve this goal, we must reconcile the different characteristics of each abstraction level.
Specifically, computational graph abstractions favor pure operators that return a new tensor for each operation. This allows us to organize computations through directed acyclic graphs and perform effective graph rewriting without worrying about side effects. On the other hand, most tensor programs and libraries of low-level computations
adopt \emph{destination-passing style}~\cite{shaikha2017dps} (DPS) interfaces, which take the computation result tensors as inputs and directly mutate them, rather than allocating and returning new tensors.
For these cases, we introduce a requirement in \sys{} to pass input and output memory explicitly to low-level tensor programs, conforming to the DPS. The DPS abstracts away memory management concerns from low-level code, thereby simplifying code generation.

We introduce two foreign function call primitives as shown in \autoref{fig:approach:cross-level-example} to bridge abstraction levels.
First, we introduce 
\calltir{}, which allows for direct invocation to a tensor program from
the graph level. 
We design the semantics of \calltir{}~(\autoref{fig:approach:calltir-semantics}) to directly map to a DPS call of the low-level function.
This approach allows us to assign high-level semantics to \calltir{}s during graph-level transformations and lower them to DPS calls with memory management later.

Notably, \calltir{} also takes the shape annotation of the output as well as potentially other symbolic expressions as arguments,
in order to pass shape information from the graph level to loop-level tensor programs.
Such shape information is crucial for optimizations on loop-level programs, like operator fusion.
By flowing the symbolic shape information
from the graph level to tensor programs, we can allow tensor programs to generate code that specializes to most static dimensions and only uses dynamic dimensions when necessary~(like dimension $n$ in \autoref{fig:approach:cross-level-example}). 

Second, we introduce the \calllibrary{} primitive to allow direct calls into
foreign operator libraries from the graph level.
In practice, it introduces great flexibility in prototyping, since external routines can be easily called from a \sys{} program. The convention of \calllibrary{} mirrors those of \calltir{}, except that the callee is instead the name of a library function. These functions are supplied by a 
registry
and linked to the final runnable module.

It is worth noting that we closely combine graphs, tensor programs
and libraries,
all of which are critical in ML compilers,
and meanwhile we leverage and complement other lower layers in compilers for
GPU source code generation.

\begin{figure}[t]
    \centering
    \vspace{-0.5em}
    \includegraphics[width=0.45\textwidth]{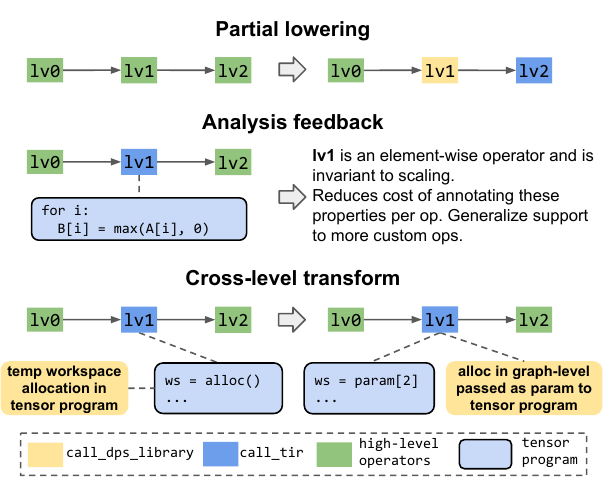}
    \vspacebeforecap
    \vspace{1.5em}
    \caption{Examples of common optimization patterns that leverages cross-level abstraction.} 
    \vspaceaftercap
    \label{fig:approach:cross-level-optimization-patterns}
\end{figure}

\MyPara{Benefits of cross-level abstractions.} 
\autoref{fig:approach:cross-level-optimization-patterns} summarizes common optimization patterns enabled by  cross-level abstractions:

\emph{Partial lowering:} Instead of making all lowering decisions in a single shot, a pass can make dispatch decisions or loop-level optimizations for part of the computations. This pattern is helpful,
for example, when we want to replace the lowering decision of a fused operator to different libraries; we can then pass the program to later passes to handle other operators.

\emph{Analysis feedback:} We can analyze loop patterns of tensor programs and automatically annotate their operator properties. Typically, compiler developers need to manually annotate properties for each individual high-level operator in the system. By adopting cross-level abstraction and instead relying on analysis-based properties, we can greatly
reduce the engineering cost of annotation on high-level operators.

\emph{Cross-level transforms:} Sometimes, optimization opportunities can only be disclosed after some low-level optimizations. 
For example, tensor program analysis
may decide that a tensor program needs a temporary workspace.
In this case, we can jointly transform the tensor program and graph level to insert the workspace allocation at graph level, allowing the workspace to also
participate as part of global memory planning.

While each optimization pattern is useful on its own, the real benefit emerges when we combine them. 
For example, we can perform partial lowering to libraries and then optimize the remaining components using other techniques.
We will discuss cross-level optimizations in detail in \S\ref{sec:composable-optimizations}.

%% file: sections/optimizations.tex
\vspace{-3pt}
\section{Cross-Level Algorithms and Optimizations}
\label{sec:composable-optimizations}

This section describes a concrete set of algorithms and optimizations that make use
of the proposed cross-level abstractions in an end-to-end compilation framework.

\subsection{Shape Annotation Deduction}
\label{sec:shape-annotation-deduction}

The symbolic shapes in annotations serve as an important source of information for optimization passes. 
To maximize the use of this information, \sys{}
automatically tracks and deduces symbolic shape annotations of
intermediate values not only during model construction but also
between compiler passes, allowing passes to 
deduce equalities and relations between shapes and enables extra optimizations. 
Meanwhile, this also increases the demand for deduction efficiency, as the deduction runs for every pass.

Each tensor operator has a registered shape deduction rule
that takes the inputs' shape annotations and values~(such as the case of \texttt{reshape})
 and returns
the output annotations.
We adopt a forward deduction method that deduces the annotation of an expression based on its inputs.
For foreign function call primitives \calltir{} and \calllibrary{},
the output annotations are part of their arguments~(\autoref{fig:approach:cross-level-example}) and will be directly used
for deduction.
Forward deduction
offers the benefits of simplicity and locality by effectively avoiding synchronization complexities across global contexts during processing.
The forward deduction also leverages the explicit information of \matchcast{}.
With forward deduction, shape annotations of any new variables introduced during compiler passes can be efficiently deduced locally.

\begin{figure}[!t]
    \centering
    \includegraphics[width=0.48\textwidth]{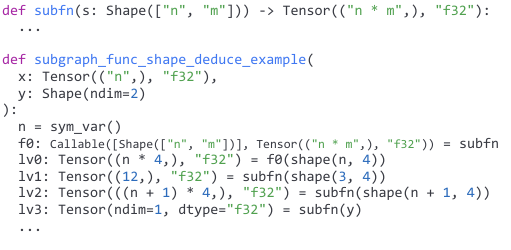}
    \vspacebeforecap
    \vspace{-0.5em}
    \caption{
        Exemplary scenarios of dynamic shape
        deduction that involve subgraph function calls.
        \texttt{subfn} contains a signature that takes shape with
        symbolic variable $n$, $m$ and returns  an one-dimensional Tensor
        that contains shape $n*m$. The annotations in \texttt{f0}, \texttt{lv0-lv3}  
        are deduced by the system.
    }
    \vspaceaftercap
    \label{fig:approach:deduction-with-cross-function}
\end{figure}

In addition, driven by the goal of providing as much shape information as possible,
we also recognize the importance of propagating interprocedural shape relations globally across subgraph function
calls to accommodate intermediate results of optimization passes like fusion, where the subgraph functions themselves contain dynamic inputs and outputs. 
\autoref{fig:approach:deduction-with-cross-function} exemplifies symbolic shape deduction for subgraph function calls. 
Our system is able to take the symbolic relations at function \texttt{subfn} and propagate the shapes correctly in the callers.
We summarize the key design principles of our shape annotation and deduction as follows:

\MyPara{Isolated symbolic relations at function boundaries.}
The function signatures in \sys{} give parameter and return value annotations, allowing for the annotation inference of function calls with only the function signature.
This allows for functions to be used as first-class values with the \texttt{Callable} annotation.
For example, we can deduce the return shapes of \texttt{subfn} calls in \autoref{fig:approach:deduction-with-cross-function} by only looking at its signature.
The function signature also serves as source to generate runtime checks, namely to ensure that passed arguments and return value match the annotations in signature.
These shape checks are lightweight and do not impact the
overall performance.

\begin{figure}[!t]
    \centering
    \vspace{-1em}
    \includegraphics[width=0.48\textwidth]{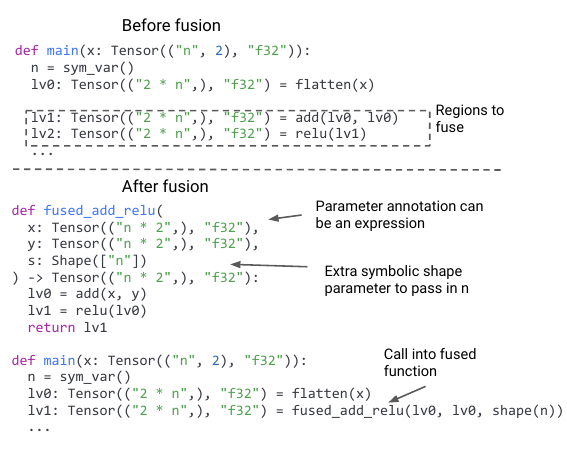}
    \vspacebeforecap
    \vspace{-0.5em}
    \caption{
        Example function signature that contains symbolic expressions caused by a result of operator fusion.
    }
    \vspaceaftercap
    \label{fig:approach:fuse-with-symbolic-args}
\end{figure}

\MyPara{Forward symbolic deduction.}
\sys{} performs forward shape deduction based
on the symbolic shape relations.
As a \emph{safety net}, coarse-grained annotations are returned when more specific information cannot be inferred (such as for data-dependent operators), as it allows the symbolic deduction to succeed for common cases but also supports 
general cases.
Note that it is permitted to pass
values with coarse-grained annotations~(e.g., \texttt{Shape(ndim=2)})
to functions that contain more specific 
annotations like \texttt{Shape((n,m))}, since we have lightweight runtime checks at the function boundary ensure the shape matches.
We choose forward deduction by default to maintain the efficiency of the symbolic shape
deductions across passes
(a full-graph forward deduction takes time linear to the number of operations),
and meanwhile still support the introduction of more powerful but less efficient deduction methods via compiler passes as needed.

\MyPara{Support symbolic expressions in parameter annotations.} Besides symbolic variables, we support general arithmetic expressions in function
parameter annotations, which is an important capability to simplify operator fusion
and other transformations across subgraph functions in
dynamic shape settings.
\autoref{fig:approach:fuse-with-symbolic-args} provides an example 
of such a case.
This example intends to fuse two intermediate operators, \texttt{add} and \texttt{relu}. 
However, both intermediate values contain an expression $2 \times n$. A na\"ive approach would create a new function with parameters \texttt{x} and \texttt{y} that have shape $(2n,)$,
but $n$ is nowhere supplied. To address this problem, the operator fusion pass passes an extra parameter $s$ which is bound to the runtime value of $n$. Passing extra symbolic arguments is a common pattern we use when designing passes that lift out function regions and recombine.

These three principles strike a balance between the need for symbolic information and the deduction complexity. They make global symbolic shape tracking feasible for common machine learning use cases and create opportunities for dynamic shape--aware optimizations.

\begin{figure*}[t]
    \centering
    \includegraphics[width=0.93\textwidth]{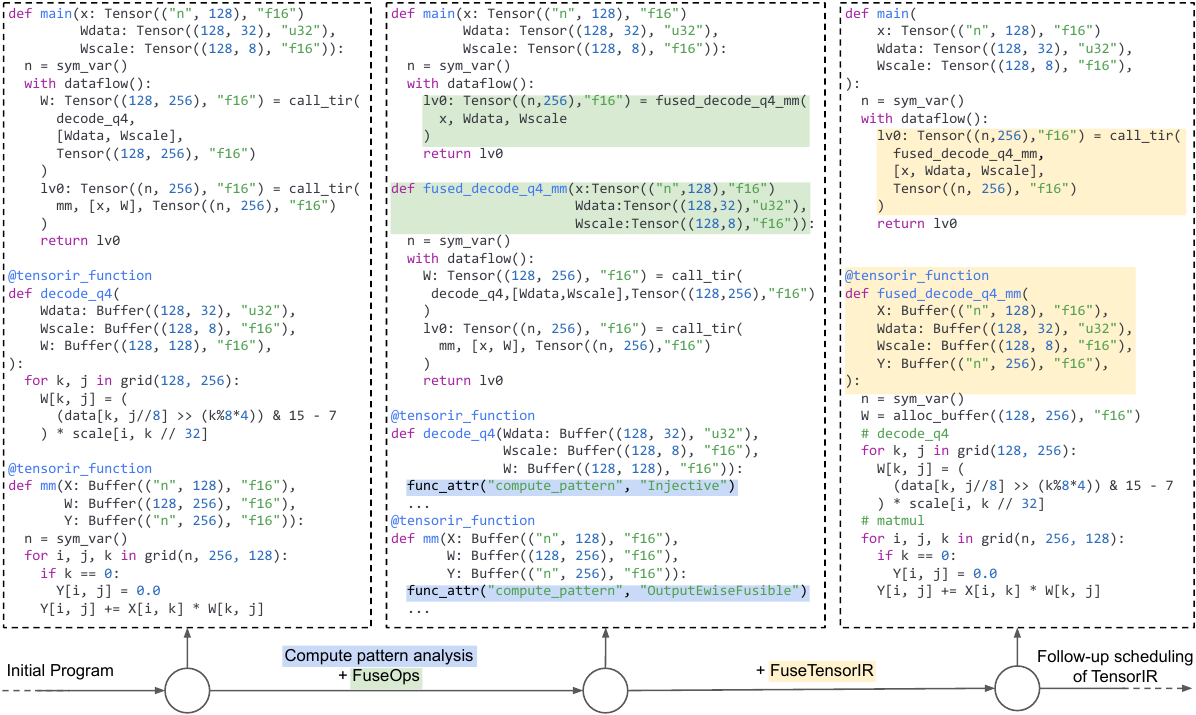}
    \caption{Dynamic shape--aware operator fusion case study with customized quantization decode (\texttt{decode\_q4}).
    Compute pattern analysis 
    classifies each tensor program 
    to a pattern kind (blue).
    Pattern-match-based
    \texttt{FuseOps}
    makes use of these
    pattern kinds to
    construct new subgraph functions
    and generate subgraph function calls (green).
    \texttt{FuseTensorIR} merges tensor programs
    and replaces subgraph function calls by \calltir{} (yellow).
    Notably, cross-level abstractions in \sys{} allows
    fusion for customized tensor programs
    that cannot be easily represented on graph level without introducing
    specialized operators,
    which is essential for browser and mobile deployment.}
    \label{fig:optim:operator-fusion}
    \vspaceaftercap
\end{figure*}

\begin{algorithm}[tb]
   \scriptsize
   \caption{Analysis feedback pass for tensor program pattern kind in \sys{}}
   \label{alg:pattern-kind-analysis}

\begin{algorithmic}[1]
   \STATE {\bfseries Input:} Tensor program function \texttt{f}.
   \STATE {\bfseries Output:} The pattern kind \texttt{kind} of tensor program \texttt{f}.
   \STATE Collect the tensor read indices \texttt{r\_indices} and write indices
   \texttt{w\_indices} in \texttt{f}.

   For example, for \texttt{C[i,j] = A[i,j] + B[j]},
   \texttt{r\_indices} is \texttt{([i,j],[j])} and
   \texttt{w\_indices} is \texttt{([i,j],)}.
   \IF{not all write indices in \texttt{w\_indices} are the same}
     \RETURN \texttt{Opaque}
   \ENDIF
   \STATE \texttt{kind} $\leftarrow$ \texttt{Opaque}
   \STATE \texttt{has\_elem\_wise} $\leftarrow$ \texttt{False}
   \FORALL{read indices \texttt{r\_idx} in \texttt{r\_indices} and the only write indices \texttt{w\_idx}}
      \IF{\texttt{IsElementWise(r\_idx,w\_idx})}
         \STATE \texttt{kind} $\leftarrow$ \texttt{ElementWise}
         \textit{(e.g., read \texttt{A[i,j]} and write \texttt{C[i,j]})}
         \STATE \texttt{has\_elem\_wise} $\leftarrow$ \texttt{True}
      \ELSIF{\texttt{IsBroadcast(r\_idx,w\_idx})}
         \STATE \texttt{kind} $\leftarrow$ \texttt{Broadcast}
         \textit{(e.g., read \texttt{B[j]} and write \texttt{C[i,j]})}
      \ELSIF{\texttt{IsInjective(r\_idx,w\_idx})}
         \STATE \texttt{kind} $\leftarrow$ \texttt{Injective}
         \textit{(e.g., read \texttt{A[j,i]} and write \texttt{C[i,j]})}
      \ENDIF
   \ENDFOR
   \IF{\texttt{kind==Broadcast} and \texttt{has\_elem\_wise==True}}
      \STATE \texttt{kind} $\leftarrow$ \texttt{ElementWise}
      \textit{(to handle cases \texttt{C[i,j]=A[i,j]+B[j]})}
   \ELSIF{\texttt{kind==Opaque} and \texttt{IsFuseMultiplyAdd(f)}}
      \STATE \texttt{kind} $\leftarrow$ \texttt{OutputEwiseFusible}
      \textit{(to handle matmul, convolution, etc.)}
   \ELSIF{\texttt{kind==Opaque} and \texttt{HasReductionLoop(f)}}
      \STATE \texttt{kind} $\leftarrow$ \texttt{Reduction}
      \textit{(to handle general reductions such as \texttt{sum} and \texttt{max})}
   \ENDIF
   \RETURN \texttt{kind}
\end{algorithmic}
   \normalsize
\end{algorithm}

\begin{algorithm}[tb]
   \scriptsize
   \caption{\texttt{FuseOps} in \sys{}}
   \label{alg:fuse-ops}

\begin{algorithmic}[1]
   \STATE {\bfseries Input:} IR module \texttt{mod} and fusion patterns \texttt{patterns}.
   \STATE {\bfseries Output:} The updated IR module after \texttt{FuseOps}.
   \STATE \texttt{mod\_updated} $\leftarrow$ \texttt{mod.Copy()}
   \FORALL{graph-level function \texttt{g} in \texttt{mod}}
      \FORALL{fusion pattern \texttt{pattern} in \texttt{patterns}}
         \STATE \texttt{matches} $\leftarrow$ \texttt{CrossLevelPatternMatch(g, mod, pattern)}
         \FORALL{\texttt{match} in \texttt{matches}}
            \STATE \texttt{subgraph\_fn} $\leftarrow$ \texttt{NewFuncWithSymShapePreserved(match)}
            \STATE \texttt{mod\_updated.AddFunc(subgraph\_fn)}
            \STATE Replace \texttt{match} in \texttt{g} with a new subgraph function call to \texttt{subgraph\_fn}.
         \ENDFOR
      \ENDFOR
   \ENDFOR
   \RETURN \texttt{mod\_updated}
\end{algorithmic}
   \normalsize
\end{algorithm}

\subsection{Cross-Level Dynamic Shape--Aware Operator Fusion}
\label{sec:optim:fusion}

Operator fusion helps to bring multiple operators
together and reduce the overall memory loading
cost of operators.
\autoref{fig:optim:operator-fusion} shows the general flow of operator fusion in \sys{}.
A \sys{} program can contain
tensor programs for both standard (e.g., matmul) and customized operators
that may not have corresponding graph-level operators,
such as
quantization decode
written in loops.
We first get an \emph{analysis feedback} pass to annotate
the pattern kind of each tensor program
by pattern matching on tensor programs
(Algorithm~\ref{alg:pattern-kind-analysis} shows the simplified pseudocode).
The candidate pattern kinds include \texttt{ElementWise},
\texttt{Broadcast}, \texttt{Injective}, \texttt{Reduction},
\texttt{OutputEwiseFusible}, or \texttt{Opaque} (as the fallback).
These pattern kinds describe the mathematical properties of tensor programs, and such information is
then
used by the \texttt{FuseOps} pass (Algorithm~\ref{alg:fuse-ops}) to group tensor program calls
into subgraph functions via pattern-match-based graph partitioning
(an example pattern is the fusion of \texttt{ElementWise} tensor programs into the back of \texttt{OutputEwiseFusible} ones, such as \texttt{matmul}+\texttt{ReLU}).
Finally, we apply \texttt{FuseTensorIR},
a \emph{cross-level transformation} that jointly
updates tensor programs and the 
graph-level calling site
by merging tensor programs
called in each subgraph function
into a single one.
Unlike applying na\"ive static-shape fusion, we
need to make sure that
all steps above handle symbolic shapes
by tracking the symbolic variables and generate
extra symbolic variable parameters to support symbolic expressions in parameter annotations (similar to \autoref{fig:approach:fuse-with-symbolic-args}), 
as we discussed in \S\ref{sec:shape-annotation-deduction}.
Notably, the analysis feedback significantly reduces manual effort, as we can automate the tensor program pattern analysis with a lightweight pass, whereas traditional single-shot abstractions require heavy and inflexible manual operator annotations.

The three fusion sub-steps bring great opportunities for further composition and customization.
For example, we can apply a pass to fuse new sets of patterns that are not
covered by \texttt{FuseOps}~(e.g., fusing all sub-operators in
scaled dot-product attention~\cite{NIPS2017_3f5ee243}),
and use \texttt{FuseOps} for the remainder. 
\texttt{FuseTensorIR} can then transform the
fused subgraph function
from both customized and standard fusion.
This approach allows quick composition of
different fusion, improving the overall productivity of continuous compiler development.

\begin{figure}[!t]
    \centering
    \includegraphics[width=0.48\textwidth]{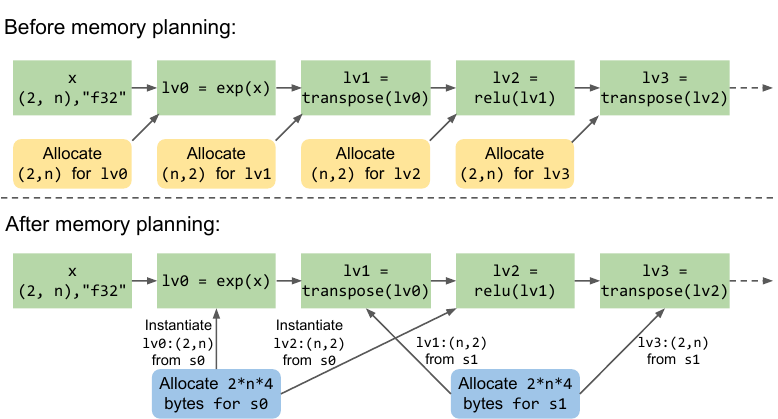}
    \vspacebeforecap
    \vspace{0.5em}
    \caption{Dynamic shape--aware memory planning example. Before planning, all \textit{four} intermediate tensors are individually allocated. After memory planning, the intermediate tensors reuse \textit{two} allocated storage chunks.}
    \label{fig:optim:memory-planning}
\end{figure}

\begin{algorithm}[tb]
   \scriptsize
   \caption{Dynamic shape--aware memory planning}
   \label{alg:memplan}

\begin{algorithmic}[1]
   \STATE {\bfseries Input:} Graph-level function \texttt{g}.
   \STATE {\bfseries Output:} The updated function after memory planning.
   \STATE Lower \calltir{} and \calllibrary{} in \texttt{g}, expanding them to explicit memory allocation and DPS calls.
   \STATE \texttt{liveness} $\leftarrow$ \texttt{LivenessAnalysis(g)}
   \textit{(for the live range of each tensor)}
   \STATE \texttt{storage\_pool} $\leftarrow$ a new storage pool with symbolic shape awareness
   \FORALL{operation \texttt{op} in \texttt{g} in the sequential order}
      \IF{\texttt{op} is a tensor allocation of tensor \texttt{t}}
         \STATE \texttt{storage} $\leftarrow$ 
         
         \texttt{\ \ storage\_pool.RequestReuseWithSymShape(op.shape,op.dtype)}
         \IF{\texttt{storage} is null}
            \STATE \texttt{storage} $\leftarrow$ \texttt{storage\_pool.NewStorage()}
            \STATE Insert the storage allocation in front of \texttt{op} in \texttt{g}.
         \ENDIF
         \STATE Insert tensor instantiation from \texttt{storage} to \texttt{t} in front of \texttt{op} in \texttt{g}.
      \ELSE
         \FORALL{tensor \texttt{t} which is used by \texttt{op}}
            \IF{\texttt{liveness.TensorIsDeadAfterOp(t, op)}}
               \STATE \texttt{storage\_pool.RecycleStorageForTensor(t)}
            \ENDIF
         \ENDFOR
      \ENDIF
   \ENDFOR
   \RETURN \texttt{g}
\end{algorithmic}
   \normalsize
\end{algorithm}

\subsection{Dynamic Shape--Aware Memory Planning}
\label{sec:optim:memplan}

Memory is an essential resource in modern ML applications.
Most ML compilers can plan memory reuse by comparing sizes of static-shape tensors and allocating a fixed set of memory blocks ahead of time to reduce the runtime memory allocation. Normally, compilers cannot take the same approach for compile-time unknown shapes and must rely on runtime memory allocators. With symbolic shape abstractions, however, we can analyze and compare dynamic tensor shapes and plan for their reuse accordingly. 
\autoref{fig:optim:memory-planning} and Algorithm~\ref{alg:memplan} show
how we apply memory planning with dynamic shapes.
We first lower the foreign function calls \calltir{} and \calllibrary{}
to explicit memory allocation and
DPS calls~(as in \autoref{fig:approach:calltir-semantics}),
so we can expose these allocation for planning.
To support dynamic shape,
the \texttt{RequestReuseWithSymShape} in Algorithm~\ref{alg:memplan}
leverages symbolic expression analysis to prove whether two
symbolic expressions are equal, and then appropriately pick
an reusable storage from the pool.
We can further
take the upper bound of the
symbolic values when they are known (e.g., annotated by users, such as the inherent context lengths in LLMs)
and statically allocate adequate memory,
which allows creating a static memory allocation plan ahead of time, even in the presence of dynamic shapes.
This upper bound approach is related to \texttt{Func::bound()}
in Halide~\cite{ragan2013halide} which marks bounds in tensor programs,
and we bring it both graph and tensor program levels.
Such predictable memory usage estimation
is crucial for deploying dynamic ML models on memory-limited backends.
Additionally, static planning can improve model performance by enabling CUDA Graph offloading~(\S\ref{sec:optim:cuda-graph}), which
relies on static memory allocation.

\subsection{Cross-Level Tensor Program Workspace Lifting}
\label{sec:optim:workspace-lift}

\begin{figure}[!t]
    \centering
    \includegraphics[width=0.40\textwidth]{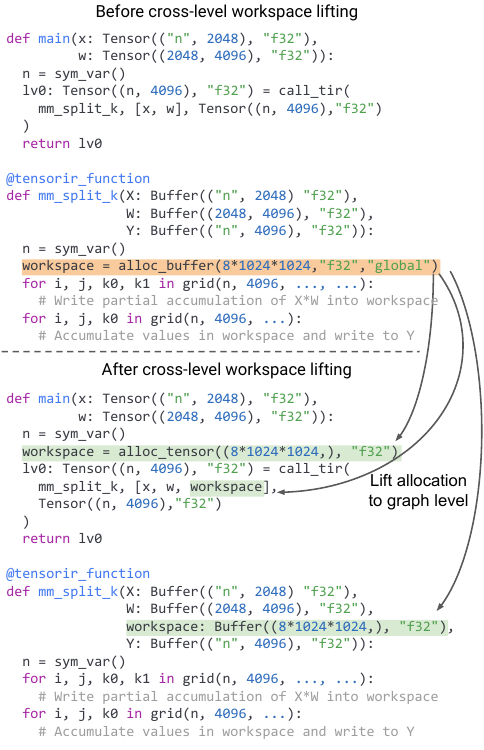}
    \caption{Cross-level tensor program workspace lifting example.
    The global memory allocation in the tensor program (orange) is lifted to
    the graph level, being passed explicitly via \calltir{}.
    The tensor program is updated accordingly
    to take in this workspace as a parameter (green).}
    \label{fig:optim:workspace-lift}
    \vspaceaftercap
\end{figure}

In addition to memory allocation at the graph level,
sometimes a tensor program may also require global memory allocation
for intermediate workspace.
For example, the Stream-K~\cite{osama2023streamkworkcentricparalleldecomposition}
schedule of \texttt{matmul} decomposes a matrix multiplication into two
phases of reduction.
The first phase writes partial accumulation results into an
intermediate global memory buffer,
which is then consumed and accumulated by the second phase.
As shown in \autoref{fig:optim:workspace-lift},
We can detect such global memory allocation in tensor programs
from \emph{analysis feedback},
and jointly rewrite the tensor program and the graph-level caller site
to lift the allocation to graph level.
Importantly, the lifted allocation can be planned by the
memory planning in \S\ref{sec:optim:memplan},
which further increases the overall memory reuse.
This optimization is only possible with the cross-level abstractions
when the shape relation is preserved throughout all
cross-level transformations in \sys{},
and the optimization opportunities for planning such memory reuse may
not arise in the traditional single-shot lowering flow.

\subsection{CUDA Graph Offloading}
\label{sec:optim:cuda-graph}

CUDA Graph~\cite{CUDAGraph} is an optimization that reduces GPU kernel launch overhead at the GPU driver level and thereby improve the overall system performance.
It works by capturing multiple GPU kernel launches and then replaying them as a group, rather than launching individual kernels separately.
To replay captured kernel launches, the GPU driver requires the all global memory accessed by the kernels to be constant-sized and statically allocated ahead of capturing. This poses a significant challenge for applying CUDA Graph to dynamic-shape ML models.
With static memory planning, \sys{} can pre-allocate all memory statically, even for dynamic-shaped tensors.
We build a pass
to analyze the computational graph and lift subgraphs that meet CUDA Graph conditions into subgraph function after memory planning.
The pass inserts runtime builtin functions that handle CUDA Graph capture or replay for these offloaded subgraph functions.
At runtime, only the first run of a subgraph function triggers
CUDA Graph capture; subsequent runs automatically replay the captured CUDA Graph. Through these, we extend CUDA Graph, typically available only for static models, to a broad range of dynamic workloads.
We can generally apply the principle of the graph offloading optimization to any GPU backend that supports static execution graphs in the future.

\begin{figure}[t]
    \centering
    \includegraphics[width=0.45\textwidth]{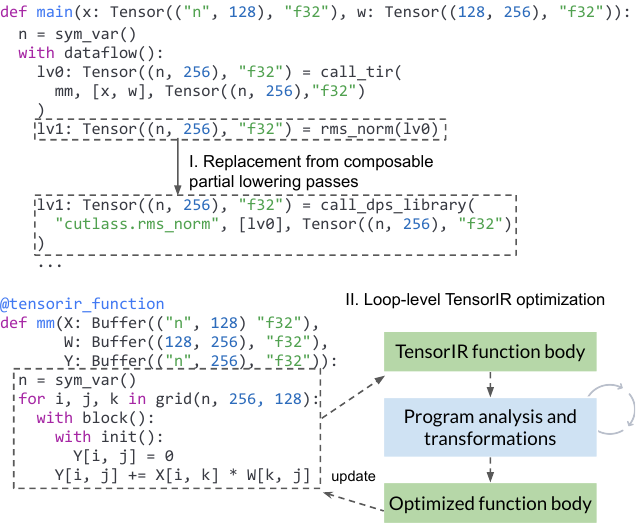}
    \caption{Tensor operator optimization examples.}
    \label{fig:optim:library-dispatch-and-tir-optim}
\end{figure}

\begin{figure}[t]
    \centering
    \includegraphics[width=0.48\textwidth]{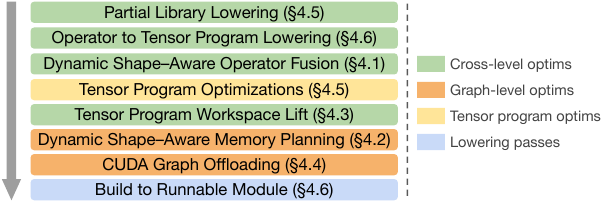}
    \vspacebeforecap
    \vspace{1em}
    \caption{Cross-level optimization and lowering pipeline.}
    \vspaceaftercap
    \label{fig:optim:pipeline}
\end{figure}

\begin{figure*}[!t]
    \centering
    \vspace{-0.5em}
    \includegraphics[width=0.95\textwidth]{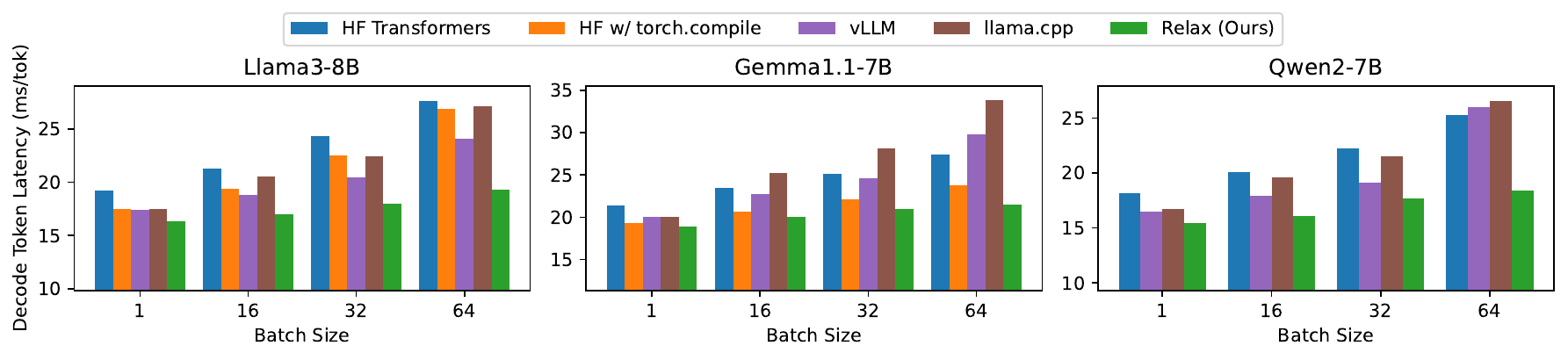}
    \vspace{-1em}
    \caption{Inference performance of various models on NVIDIA RTX~4090 under different batch sizes.
    We omitted the results of HF with \texttt{torch.compile} for Qwen2-7B due to the lack of support.
    \sys{} brings competitive performance across different models and batch sizes, and reduces the decode token latency by up to 27\%.}
    \vspace{-1em}
    \label{fig:llm-4090}
\end{figure*}

\subsection{Tensor Operator Optimizations via Partial Lowering}
\label{sec:optim:partial-lowering}

Modern ML frameworks optimize tensor operators in two approaches.
We can offload computation to platform-specific operator libraries,
or leverage compiler tensor program optimizations and code generation.
Most existing ML compilers make
these decisions at the boundary between the graph level and lower levels,
making it hard to compose different lowering approaches.
For instance, if we want to introduce a new operator library, we need to carefully examine the existing lowering strategies and update accordingly.
The complexity of the lowering layer grows as we
incorporate more approaches, such as different ways
of auto-scheduling. 

\sys{} applies tensor operator optimizations via \emph{partial lowering}~(\autoref{fig:optim:library-dispatch-and-tir-optim}).
We 
register a set of ``(subgraph pattern, library function)'' pairs in
\sys{}, and
build pattern-match-and-rewrite passes that detect specific patterns
(e.g., matmul with epilogue)
in the graph level, and partially lower
detected regions to foreign library calls.
\sys{} also allows users to register patterns for customizability.
Additionally,
based on TensorIR~\cite{feng2023tensorir} scheduling transformations,
we build a set of analysis-based dynamic shape--aware 
schedule rules to optimize
tensor programs
by minimizing memory loading.
We 
can
also include
passes to apply
Ansor-style~\cite{chen2019learning,ansor,shao2022tensor}
auto-tuning for rare tensor programs (e.g., complicated convolutions)
that our analysis-based schedule rules fail to handle.
Importantly,
all these transformations can be composed
together and work collectively.
\sys{} enables fast development and only requires a single relatively simple partial lowering pass to customize the optimizations
of
operators.

\subsection{Optimization and Lowering Pipeline}

\sys{} uses a fixed-order pipeline
(without fixed point) on the cross-level abstraction
to optimize, lower and finally build an end-to-end model
into a runnable module.
An example pipeline is shown in \autoref{fig:optim:pipeline}.
We prioritize the partial library lowering~(\S\ref{sec:optim:partial-lowering})
to leverage external library functions on the target platform.
Next, we go through the whole program, generate tensor programs for
all high-level operator calls, and lower the operator calls
to \calltir{} of corresponding tensor programs.
We can then apply operator fusion,
tensor program workspace lifting,
memory planning,
and CUDA Graph offloading.
Notably, some tensor program optimizations (e.g., workspace lifting)
are applied before graph optimizations,
which necessitates \sys{}'s cross-level abstraction design.

At the end of the pipeline is building the model into a runnable module.
At the graph level,
a fundamental task is to associate symbolic variables with concrete shape values and compute symbolic expressions at runtime.
We create an integer host tensor to store runtime values of all symbolic expressions in the program.
At the start of transformation, we populate the values of symbolic variables in program input tensors. We then generate tensor programs that load from the tensor, evaluate symbolic expressions, and store results to corresponding locations. Finally, we insert function calls to construct shape tuples when tensor shapes are needed as first-class values.
After this transformation, we erase all annotations, leaving a program comprised mainly of low-level function calls.
The calls will then be translated
to a sequence of virtual machine instructions, each of which is a call into a generated or builtin function.
For optimized low-level tensor programs, we directly generate
corresponding GPU code.
We package the graph-level virtual machine instructions and the GPU code together into a single holistic end-to-end module, which can then run on the target platform of compilation.

%% file: sections/evaluation.tex
\section{Evaluation}
\label{sec:evaluation}

We implement \sys{} on top of Apache TVM~\cite{chen2018tvm}. Notably, the insights presented in this paper can also benefit other ML compilation frameworks as well. This section provides evaluation to answer the following questions:

\squishlist
    \vspace{-0.2em}
    \item
    Does \sys{} provide competitive LLM inference performance compared with existing frameworks?
    (\S\ref{sec:eval:llm-traditional-gpus})?
    \vspace{-0.2em}
    \item What
    are the effects of proposed abstractions and
    optimizations on performance and memory usage
    (\S\ref{sec:eval:ablation})?
    \vspace{-0.2em}
    \item Can \sys{} support these emerging LLMs on a broad set of emerging platforms
    (\S\ref{sec:eval:more-emerging-platforms})?
    \vspace{-0.2em}
    \item How does \sys{}\
    perform on a broader model set?
    (\S\ref{sec:eval:more-modality})?
\squishend

\subsection{Large Language Model Inference Evaluation}
\label{sec:eval:llm-traditional-gpus}

\begin{figure*}[!t]
    \centering
    \includegraphics[width=0.95\textwidth]{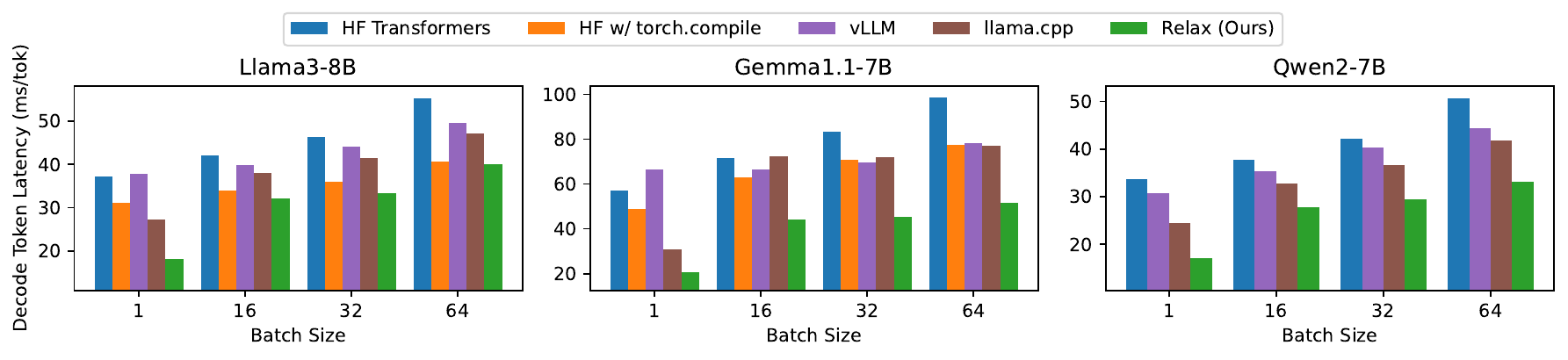}
    \vspace{-1em}
    \caption{Inference performance of various models on AMD Radeon 7900~XTX under different batch sizes.
    \sys{} consistently delivers competitive performance,
    and brings optimized performance with up to 1.50x under case of batch size 1.}
    \vspace{-1em}
    \label{fig:llm-7900}
\end{figure*}

\begin{figure*}[!t]
    \centering
    \includegraphics[width=0.95\textwidth]{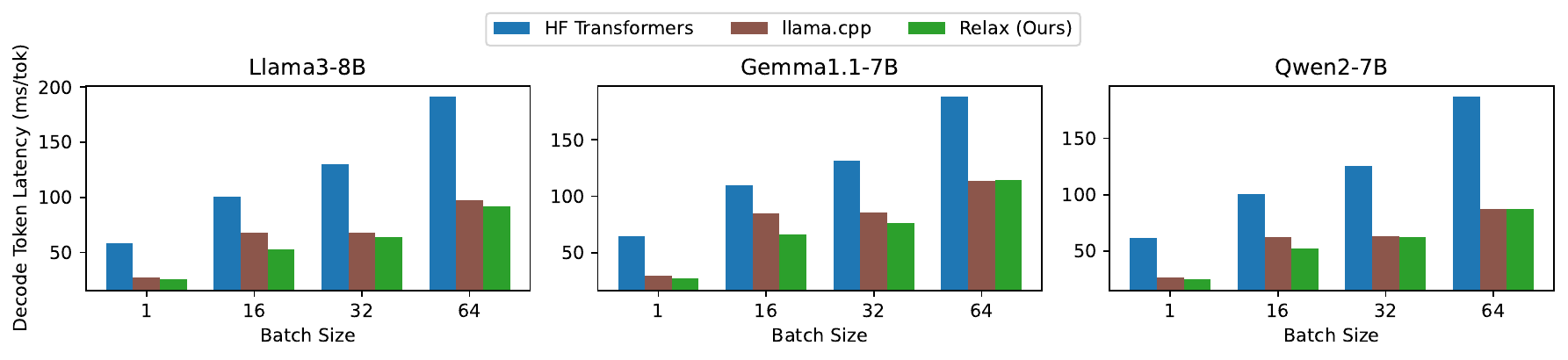}
    \vspace{-1em}
    \caption{Inference performance of various models on Apple M2~Ultra under different batch sizes.
    \sys{} has competitive performance comparing to the hand-optimized llama.cpp baseline.
    }
    \label{fig:llm-m2ultra}
\end{figure*}

This section evaluates the performance of \sys{} with end-to-end LLMs, a typical class of emerging ML models, on NVIDIA GPUs and emerging AMD and Apple GPUs.
We assess the per-token latency of the LLM generation decode phase across different batch sizes,
so that dynamism of both sequence length and batch size is included.
Our evaluation is conducted on Llama3-8B, Gemma1.1-7B and Qwen2-7B with \texttt{float16} weight and activations with NVIDIA RTX~4090, AMD Radeon 7900~XTX and Apple M2~Ultra.
We compare with baseline frameworks HuggingFace Transformers (v4.41.2)~\cite{hf_transformers} with PyTorch (v2.3.1)
eager~\cite{paszke2019pytorch} and compile mode~\cite{pytorch2}, vLLM~\cite{kwon2023efficient} (v0.5.0.post1), and hand-optimized LLM inference system \texttt{llama.cpp}~(\texttt{172c825})~\cite{llama.cpp}.
FlashAttention~\cite{flashattn} is enabled for baselines when available.
We construct \sys{} IR 
with a PyTorch-like \texttt{nn.Module} interface.
We measure the decode time of generating 32 tokens
and compute the per-token latency per sequence.
Importantly, \sys{} compiles models only once for arbitrary batch sizes and sequence lengths.

\autoref{fig:llm-4090} to \ref{fig:llm-m2ultra}
show that \sys{} brings consistently competitive performance 
across platforms.
In these cases, symbolic shape analyses enable \sys{} to generate tensor programs that
are only dynamic for the batch size and sequence length dimensions.
Cross-level abstractions allow seamless composition of graph and tensor program optimizations
for any GPU backend, eliminating the needs to manually write kernels for each individual backend.
More importantly, cross-level abstractions enable us to use compiler-optimized matrix-vector multiplication tensor programs at batch size 1, while being able to apply partial library lowering to leverage operator libraries for other batch sizes.
This 
flexibility greatly helps improve the performance of special cases.

Notably, while the HuggingFace Transformers with 
PyTorch 
compile mode supports dynamic sequence lengths, it still requires static KV cache, which depends on significant model definition changes and is therefore available only for a few models.\footnote{\url{https://huggingface.co/docs/transformers/main/en/llm_optims}}
Additionally, not every baseline well supports all the platforms.
The hand-optimized llama.cpp has strong performance on Apple GPUs but performs less effectively on NVIDIA GPUs.
And PyTorch compile mode and vLLM lack Apple GPU support.
In contrast, \sys{} is able to support all the platforms with competitive performance.

\subsection{Effects of Composable Optimizations}
\label{sec:eval:ablation}

\begin{figure}[t]
    \centering
    \includegraphics[width=0.40\textwidth]{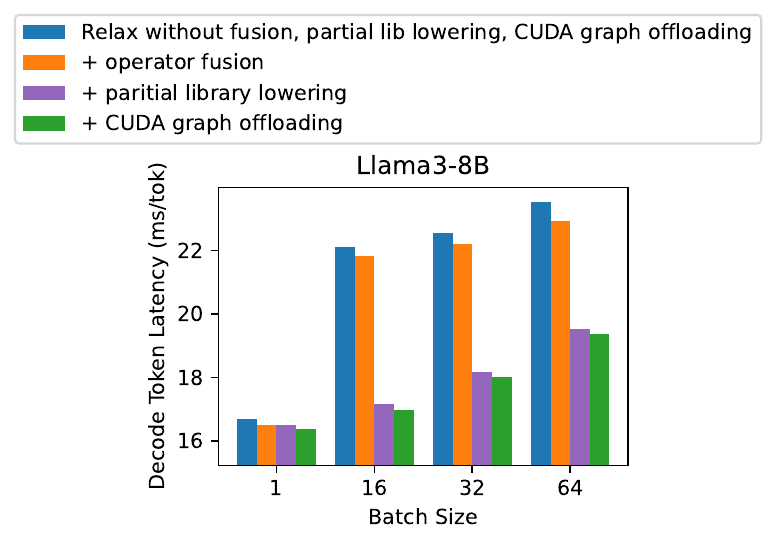}
    \vspacebeforecap
    \vspace{0.5em}

    \caption{Effects of operator fusion, partial library dispatching and CUDA Graph offloading on Llama3-8B inference across different batch sizes.}
    \vspaceaftercap
    \label{fig:llm-ablation-perf}
\end{figure}

\MyPara{Effects on performance.}
\sys{}'s 
abstraction allows for flexible combination of
optimizations such as CUDA Graph offloading, operator fusion, partial library lowering and code generation. 
We evaluate these optimizations with \texttt{float16} Llama3-8B with on NVIDIA RTX~4090.
\autoref{fig:llm-ablation-perf} shows the ablation study results.
Partial library lowering contributes the most, up to 27\% of the performance for large batch sizes, where it lowers heavy-load matrix multiplications
(which count for about 1/3 of all operators)
to cuBLAS library kernels.
Operator fusion helps by fusing
about 1/5 of all
operators (such as RMSNorm and element-wise addition)
to reduce launched kernels and GPU global memory accesses.
CUDA Graph offloading overall brings about 1-2\% of performance gain
by reducing kernel launch overheads at GPU driver level.
All these composable optimizations collectively improve the overall system performance.

\begin{table}[t]
    \small
    \centering
    \vspace{-0.5em}
    \caption{Memory usage of Llama3-8B inference with and without static memory planning, through workloads of successive prefill of different sequence lengths and successive decode of different batch sizes.}
    \vspace{-1em}
    \label{tab:eval-ablation-memory}
    \begin{tabular}{l|c}
    \toprule
    \textbf{Llama3-8B Prefill}  & MiB \\
    \midrule
    \sys{} w/o planning & 192.7 \\
    \sys{} w/. planning & 149.7 \\
    \bottomrule
    \end{tabular}
    \quad
    \begin{tabular}{l|c}
    \toprule
    \textbf{Llama3-8B Decode}  & MiB \\
    \midrule
    \sys{} w/o planning & 150.0 \\
    \sys{} w/. planning & 88.2 \\
    \bottomrule
    \end{tabular}
    \vspace{-1em}
\end{table}

\MyPara{Effects on memory usage.}
We study the memory reduction of static memory planning by measuring the total allocated activation memory size during the \texttt{float16} Llama3-8B prefill of successive inputs of length 128, 256, 512, 1024, and during the decode of successive batches of size 1, 16, 32 and 64.
The memory planning plans with the upper bound of sequence length and batch size.
When memory planning is disabled, we use a runtime memory pool to recycle unused memory.
As in \autoref{tab:eval-ablation-memory}, static memory planning reduces activation memory by 22\% during successive prefill phases
and by 40\% for decode.
With static memory planning, we always reuse memory
across all input lengths and batch sizes, even as they vary over time.
In contrast, without memory planning, the system repeatedly allocates dynamic-sized memory whenever the input shape changes,
which is unpredictable in real-world applications.
This potentially leads to even higher memory usage unless all memory is statically planned.
Additionally, by allocating all memory in advance, we enable CUDA Graph for emerging models and yield further performance gains.

Importantly,
all these optimizations rely on proposed abstractions.
For example, in the best case without shape information,
we can no longer run static analyses such as static memory planning
and static graph capture for optimizations,
which may cause additional memory and latency overhead.

\subsection{Evaluation on More Emerging Platforms}
\label{sec:eval:more-emerging-platforms}

\begin{table}[]
\small
    \centering
    \begin{talltblr}[
        caption={Inference performance~(throughput, tokens/sec) of 4-bit quantized Llama3-8B, Phi3-mini-4k and RedPajama-3B models on a broad set of emerging platforms, including mobile and embedded devices. \sys{} deploys emerging models across these platforms,
        which most existing ML frameworks do not well support.
        },
        label={table:more-emerging-platforms},
        note{$\dag$}={We run 3-bit quantized Llama2-7B on iPhone and 4-bit quantized Llama2-7B on Samsung S23 to fit the VRAM limit of the mobile environments.},
    ]{colspec={Q[l] Q[l] | Q[c] Q[c] Q[c] }}
    \toprule
    \textbf{Devices} & \textbf{Backend} & \textbf{Llama} & \textbf{Phi3} & \textbf{RedPajama} \\
    \midrule
    iPhone 14 Pro & Metal  & 5.1\TblrNote{$\dag$}  & 13.8 & 19.5    \\
    Samsung S23   & OpenCL & 7.9\TblrNote{$\dag$}  & 13.1 & 20.5   \\
    Orange Pi 5   & OpenCL & 2.3  & 5.0  & 6.1        \\
    Steam Deck    & Vulkan & 14.0 & 20.2 & 22.9   \\
    Jetson Orin   & CUDA   & 32.0 & 59.1 & 65.2   \\
    \makecell[cc]{WebGPU \\ (M3 Max)} & WebGPU & 37.8 & 68.0 & 68.6   \\
    \bottomrule
    \end{talltblr}
    \vspaceaftercap
\end{table}

\begin{figure}[t]
    \centering
    \includegraphics[width=0.26\textwidth]{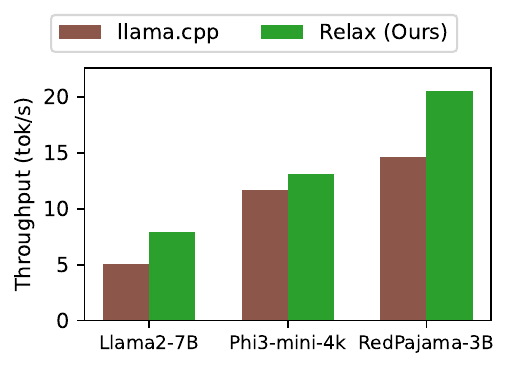}

    \caption{Single-sequence generation performance comparison of 4-bit quantized LLMs on Samsung S24. \sys{} delivers up to 55\% more throughput on evaluated models.}
    \vspaceaftercap
    \label{fig:llm-android}
\end{figure}

This section evaluates the deployment of emerging models onto a broader set of emerging platforms that are less supported by existing solutions.
We evaluate single-sequence LLM inference performance on iPhone 14 Pro with Apple A16, Samsung S23 with Qualcomm Snapdragon 8 Gen 2,
Orange Pi 5 with ARM Mali GPU, Steam Deck with AMD APU,
NVIDIA Jetson Orin developer kit,
and in-browser WebGPU~\cite{dakkak2016webgpu}
on Apple M3 Max laptop.
We run 4-bit quantized Llama3-8B for most cases, while using Llama2-7B for mobile phones
so the total memory usage fits the
VRAM limit.

As in \autoref{table:more-emerging-platforms},
\sys{} provides a
throughput
of over 5 tokens/s for mobile devices, and 2.3 tokens/s for Orange Pi 5.
In addition to physical devices, \sys{}
enables LLM deployment on the emerging 
WebGPU backend, supporting web-native machine learning.
To our best knowledge, \sys{} is the first solution to enable GPU-accelerated LLM inference on these platforms except the NVIDIA Jetson Orin.
Without memory planning that pre-allocates all needed memory 
and keeps it within the budget, these models are not even runnable on some of the environments due to memory constraints.

We further compare 
\sys{} on Samsung S23 with llama.cpp.
\sys{} achieves up to 55\% better throughput (\autoref{fig:llm-android}).
Notably, llama.cpp only utilizes CPU
due to lack of kernels for Android GPUs,
whereas \sys{} automatically generates optimized GPU codes via compilation, enabling 
emerging model deployment not only specifically on Android platforms, but also generically on more emerging platforms.

\subsection{Evaluation on Additional Set of Models}
\begin{figure}[!t]
    \centering
    \includegraphics[width=0.41\textwidth]{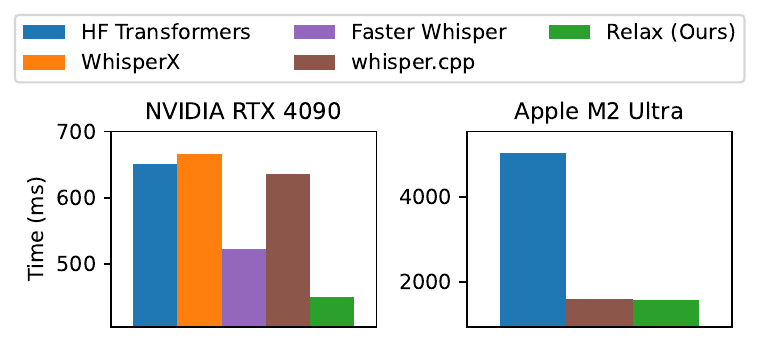}

    \caption{Transcription time of a 30-second speech file with Whisper-large-v3 on NVIDIA RTX~4090 and Apple M2~Ultra. WhisperX and Faster Whisper have no Apple GPU support. \sys{} delivers competitive performance on both NVIDIA and Apple platforms.}
    \vspaceaftercap
    \label{fig:eval-whisper}
\end{figure}
\begin{figure}[!t]
    \centering
    \includegraphics[width=0.35\textwidth]{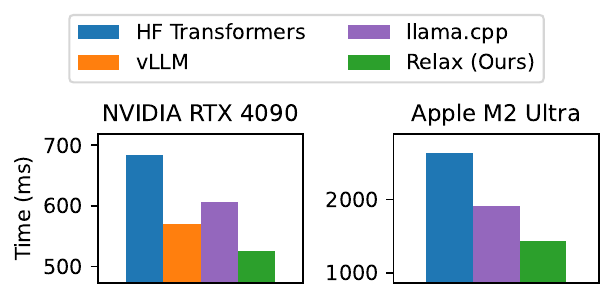}

    \caption{LLaVA generation time of 32-token for an image input on NVIDIA RTX~4090 and Apple M2~Ultra. \sys{} achieves competitive optimized performance for LLaVA generation on both platforms.}
    \vspaceaftercap
    \label{fig:eval-llava}
\end{figure}

We also study \sys{} on an additional set of models.
Whisper~\cite{radford2022robust-whisper} is an automatic speech recognition (ASR) model
implemented as an encoder-decoder Transformer.
We evaluate the time to transcribe a 30-second speech with Whisper-large-v3,
and compare \sys{} with HuggingFace Transformers, WhisperX~\cite{bain2023whisperx}~(\texttt{f2da2f8}), Faster Whisper~(v1.0.2) and whisper.cpp~\cite{whisper.cpp} (\texttt{5d950c4}).
As shown in \autoref{fig:eval-whisper},
\sys{} brings a speedup of 14\% on NVIDIA 4090 and has competitive performance on Apple GPU.
LLaVA~\cite{liu2023llava} is a large multi-modal model that integrates the pre-trained CLIP~\cite{radford2021learning-clip} visual encoder and the LLM Vicuna~\cite{vicuna2023} for general-purpose visual and language understanding.
We evaluate the time of generating 32 tokens for an image
using baselines of HuggingFace Transformers, vLLM, and llama.cpp.
Results are shown in \autoref{fig:eval-llava}. \sys{} efficiently supports the vision encoder together with the prefill and decode phases of LLM.

\label{sec:eval:more-modality}

%% file: sections/related-work.tex
\section{Related Work}
\label{sec:related-work}

Vendor-optimized libraries like cuDNN~\cite{chetlur2014cudnn}, CUTLASS~\cite{Thakkar_CUTLASS_2023}, MKL-DNN~\cite{intel2017mkldnn}, and MIOpen~\cite{jeh2019miopen} are frequently used by ML frameworks~\cite{abadi2016tensorflow, paszke2019pytorch} to support tensor operators on various hardware backends.
The libraries are platform-specific and have large engineering development costs to cover
the growing demands of operators, data formats, and layouts.
\sys{} complements such libraries by allowing them to be used alongside loop-level code optimized with dynamic shape--awareness.
Frameworks that rely on libraries can potentially use \sys{} to choose between libraries or generated code.

The emerging demand for large language models has also inspired a set of frameworks~\cite{hf_transformers, kwon2023efficient, ggml, llama.cpp} optimized for these particular workloads. 
These frameworks usually rely on manual optimizations for each specific backend.
They can leverage \sys{} to reduce the effort for supporting a broader set of workloads and emerging backends.

There has also been much work on abstractions for transforming and optimizing loop-level code for tensor operators.
Triton~\cite{tillet2019triton} and Graphene~\cite{hagedorn2023graphene}  
are abstractions that optimize tensorized programs on GPU.
DietCode~\cite{MLSYS2022_f89b79c9}, CoRA~\cite{MLSYS2022_afe8a457}, and SparseTIR~\cite{ye2023sparsetir}, 
focus on tensor program optimizations with shape dynamism and irregularity.
Mosaic~\cite{bansal2023mosaic} is a sparse compiler combining library dispatch and sparse tensor program optimizations.
Cortex~\cite{fegade2021cortex} enables tensor program optimizations for recursive computation.
We use TensorIR~\cite{feng2023tensorir} as the tensor program abstraction in our
cross-level design implementation, but we could combine our approaches with
other abstractions for
tensor supports 
programs to enable a broader spectrum of tensor program optimizations.

ML compilers are designed to represent and optimize end-to-end model computations.
High-level computations are usually represented as computation graph--style dialects.
TVM~\cite{chen2018tvm}'s Relay~\cite{Roesch_2018} and MLIR dialects~\cite{lattner2021mlir} 
represent dynamic dimensions as unknown and do not track dynamic shape relations.
IREE~\cite{iree} provides end-to-end compilation with MLIR.
Nimble~\cite{MLSYS2021_nimble} leverages runtime bucketing to support dynamic operators.
DISC~\cite{zhu2021disc,zheng2022astitch} enables shape as a first-class value but 
does not track symbolic shapes.
TorchInductor~\cite{pytorch2} brings native symbolic shape support to the PyTorch compiler,
focusing on kernel generation for TorchFX graphs~\cite{torchfx} derived from TorchDynamo~\cite{pytorch2}. 
The PyTorch compiler stores a global symbolic variable table
for traced subgraphs, and is synergistic with its JIT-focused design
and avoids cross-function symbolic shape tracking.
\sys{} complements the PyTorch compiler by abstracting and globally tracks cross-function symbolic shape for full programs, 
enabling AOT compilation and holistic deployment onto emerging platforms.
As a result, \sys{} can be used as a backend for 
frameworks like PyTorch, JAX~\cite{google-jax} to deploy models to more emerging backends.
Axon~\cite{collins2022axon} is a functional language that considers shape deduction
in its types and applies a constraint solver to determine shape relations; unlike \sys{}, it does not describe a dynamic fallback when shapes cannot be deduced statically. (Note that \sys{} could still apply a similar constraint-solving approach, despite its additional compile time costs.)
Halide~\cite{ragan2013halide} supports external function calls via \texttt{Func::define\_extern()} in tensor programs.
\sys{} extends this mechanism to both graph and tensor program levels,
bridging together external libraries and these levels.
Additionally, most existing ML compilers follow a multi-level single-shot lowering approach,
whereas \sys{} enables global symbolic shape tracking across functions via cross-level abstractions.
\sys{}'s insights for supporting dynamic shapes and cross-level optimizations can be used to improve these ML compiler frameworks.

%% file: sections/conclusion.tex
\vspace{-3pt}
\section{Conclusion}
\label{sec:conclusion}

We introduce \sys{}, an abstraction for end-to-end dynamic machine learning on emerging platforms.
Our cross-level abstractions
and first-class symbolic shapes enable
composable optimizations of dynamic shape models and allow us to
build an AOT end-to-end holistic framework that  deploys emerging models to diverse emerging backends.
\sys{} delivers performance competitive to state-of-the-art systems across platforms, including 27\% of LLM decode token latency reduction on NVIDIA GPUs.
We hope this work will encourage additional studies of dynamic shape--aware program abstractions and highlight new possibilities for ML compilers.

%% file: sections/acknowledgement.tex
\begin{acks}
We thank all anonymous ASPLOS reviewers and our shepherd Fredrik Kjolstad for their constructive feedback and comments. 
This work was supported in part by gifts from OctoAI, Qualcomm, and CMU open-source software fellowships.
\end{acks}

%% file: main.bbl
\begin{thebibliography}{10}

\bibitem{abadi2016tensorflow}
Mart{\'\i}n Abadi, Paul Barham, Jianmin Chen, Zhifeng Chen, Andy Davis, Jeffrey Dean, Matthieu Devin, Sanjay Ghemawat, Geoffrey Irving, Michael Isard, et~al.
\newblock Tensorflow: a system for large-scale machine learning.
\newblock In {\em 12th USENIX symposium on operating systems design and implementation (OSDI 16)}, pages 265--283, 2016.

\bibitem{pytorch2}
Jason Ansel, Edward Yang, Horace He, Natalia Gimelshein, Animesh Jain, Michael Voznesensky, Bin Bao, Peter Bell, David Berard, Evgeni Burovski, Geeta Chauhan, Anjali Chourdia, Will Constable, Alban Desmaison, Zachary DeVito, Elias Ellison, Will Feng, Jiong Gong, Michael Gschwind, Brian Hirsh, Sherlock Huang, Kshiteej Kalambarkar, Laurent Kirsch, Michael Lazos, Mario Lezcano, Yanbo Liang, Jason Liang, Yinghai Lu, C.~K. Luk, Bert Maher, Yunjie Pan, Christian Puhrsch, Matthias Reso, Mark Saroufim, Marcos~Yukio Siraichi, Helen Suk, Shunting Zhang, Michael Suo, Phil Tillet, Xu~Zhao, Eikan Wang, Keren Zhou, Richard Zou, Xiaodong Wang, Ajit Mathews, William Wen, Gregory Chanan, Peng Wu, and Soumith Chintala.
\newblock Pytorch 2: Faster machine learning through dynamic python bytecode transformation and graph compilation.
\newblock In {\em Proceedings of the 29th ACM International Conference on Architectural Support for Programming Languages and Operating Systems, Volume 2}, ASPLOS '24, page 929–947, New York, NY, USA, 2024. Association for Computing Machinery.

\bibitem{onnx}
Junjie Bai, Fang Lu, Ke~Zhang, et~al.
\newblock Onnx: Open neural network exchange.
\newblock \url{https://github.com/onnx/onnx}, 2019.

\bibitem{bain2023whisperx}
Max Bain, Jaesung Huh, Tengda Han, and Andrew Zisserman.
\newblock Whisperx: Time-accurate speech transcription of long-form audio, 2023.

\bibitem{bansal2023mosaic}
Manya Bansal, Olivia Hsu, Kunle Olukotun, and Fredrik Kjolstad.
\newblock Mosaic: An interoperable compiler for tensor algebra.
\newblock {\em Proceedings of the ACM on Programming Languages}, 7(PLDI):394--419, 2023.

\bibitem{black2022gptneox20b}
Sid Black, Stella Biderman, Eric Hallahan, Quentin Anthony, Leo Gao, Laurence Golding, Horace He, Connor Leahy, Kyle McDonell, Jason Phang, Michael Pieler, USVSN~Sai Prashanth, Shivanshu Purohit, Laria Reynolds, Jonathan Tow, Ben Wang, and Samuel Weinbach.
\newblock Gpt-neox-20b: An open-source autoregressive language model, 2022.

\bibitem{chen2018tvm}
Tianqi Chen, Thierry Moreau, Ziheng Jiang, Lianmin Zheng, Eddie Yan, Haichen Shen, Meghan Cowan, Leyuan Wang, Yuwei Hu, Luis Ceze, et~al.
\newblock Tvm: An automated end-to-end optimizing compiler for deep learning.
\newblock In {\em 13th USENIX Symposium on Operating Systems Design and Implementation (OSDI 18)}, pages 578--594, 2018.

\bibitem{chen2019learning}
Tianqi Chen, Lianmin Zheng, Eddie Yan, Ziheng Jiang, Thierry Moreau, Luis Ceze, Carlos Guestrin, and Arvind Krishnamurthy.
\newblock Learning to optimize tensor programs, 2019.

\bibitem{chetlur2014cudnn}
Sharan Chetlur, Cliff Woolley, Philippe Vandermersch, Jonathan Cohen, John Tran, Bryan Catanzaro, and Evan Shelhamer.
\newblock cudnn: Efficient primitives for deep learning, 2014.

\bibitem{vicuna2023}
Wei-Lin Chiang, Zhuohan Li, Zi~Lin, Ying Sheng, Zhanghao Wu, Hao Zhang, Lianmin Zheng, Siyuan Zhuang, Yonghao Zhuang, Joseph~E. Gonzalez, Ion Stoica, and Eric~P. Xing.
\newblock Vicuna: An open-source chatbot impressing gpt-4 with 90\%* chatgpt quality, March 2023.

\bibitem{collins2022axon}
Alexander Collins and Vinod Grover.
\newblock Axon: A language for dynamic shapes in deep learning graphs, 2022.

\bibitem{dakkak2016webgpu}
Abdul Dakkak, Carl Pearson, and Wen-mei Hwu.
\newblock Webgpu: A scalable online development platform for gpu programming courses.
\newblock In {\em 2016 IEEE International Parallel and Distributed Processing Symposium Workshops (IPDPSW)}, pages 942--949. IEEE, 2016.

\bibitem{flashattn}
Tri Dao, Dan Fu, Stefano Ermon, Atri Rudra, and Christopher R\'{e}.
\newblock Flashattention: Fast and memory-efficient exact attention with io-awareness.
\newblock In S.~Koyejo, S.~Mohamed, A.~Agarwal, D.~Belgrave, K.~Cho, and A.~Oh, editors, {\em Advances in Neural Information Processing Systems}, volume~35, pages 16344--16359. Curran Associates, Inc., 2022.

\bibitem{fegade2021cortex}
Pratik Fegade, Tianqi Chen, Phillip Gibbons, and Todd Mowry.
\newblock Cortex: A compiler for recursive deep learning models.
\newblock {\em Proceedings of Machine Learning and Systems}, 3:38--54, 2021.

\bibitem{MLSYS2022_afe8a457}
Pratik Fegade, Tianqi Chen, Phillip Gibbons, and Todd Mowry.
\newblock The cora tensor compiler: Compilation for ragged tensors with minimal padding.
\newblock In D.~Marculescu, Y.~Chi, and C.~Wu, editors, {\em Proceedings of Machine Learning and Systems}, volume~4, pages 721--747, 2022.

\bibitem{feng2023tensorir}
Siyuan Feng, Bohan Hou, Hongyi Jin, Wuwei Lin, Junru Shao, Ruihang Lai, Zihao Ye, Lianmin Zheng, Cody~Hao Yu, Yong Yu, et~al.
\newblock Tensorir: An abstraction for automatic tensorized program optimization.
\newblock In {\em Proceedings of the 28th ACM International Conference on Architectural Support for Programming Languages and Operating Systems, Volume 2}, pages 804--817, 2023.

\bibitem{google-jax}
Roy Frostig, Matthew Johnson, and Chris Leary.
\newblock Compiling machine learning programs via high-level tracing.
\newblock 2018.

\bibitem{ggml}
Georgi Gerganov.
\newblock ggml.
\newblock \url{https://github.com/ggerganov/ggml}, 2022.

\bibitem{whisper.cpp}
Georgi Gerganov.
\newblock whisper.cpp.
\newblock \url{https://github.com/ggerganov/whisper.cpp}, 2022.

\bibitem{llama.cpp}
Georgi Gerganov.
\newblock llama.cpp.
\newblock \url{https://github.com/ggerganov/llama.cpp}, 2023.

\bibitem{CUDAGraph}
Alan Gray.
\newblock Getting started with cuda graphs, Sep 2019.

\bibitem{hagedorn2023graphene}
Bastian Hagedorn, Bin Fan, Hanfeng Chen, Cris Cecka, Michael Garland, and Vinod Grover.
\newblock Graphene: An ir for optimized tensor computations on gpus.
\newblock In {\em Proceedings of the 28th ACM International Conference on Architectural Support for Programming Languages and Operating Systems, Volume 3}, pages 302--313, 2023.

\bibitem{intel2017mkldnn}
Intel.
\newblock Intel® math kernel library for deep learning networks, 2017.

\bibitem{iree}
{IREE Project}.
\newblock {IREE}, sep 2019.

\bibitem{jeh2019miopen}
Jehandad Khan, Paul Fultz, Artem Tamazov, Daniel Lowell, Chao Liu, Michael Melesse, Murali Nandhimandalam, Kamil Nasyrov, Ilya Perminov, Tejash Shah, Vasilii Filippov, Jing Zhang, Jing Zhou, Bragadeesh Natarajan, and Mayank Daga.
\newblock Miopen: An open source library for deep learning primitives, 2019.

\bibitem{kwon2023efficient}
Woosuk Kwon, Zhuohan Li, Siyuan Zhuang, Ying Sheng, Lianmin Zheng, Cody~Hao Yu, Joseph~E Gonzalez, Hao Zhang, and Ion Stoica.
\newblock Efficient memory management for large language model serving with pagedattention.
\newblock {\em arXiv preprint arXiv:2309.06180}, 2023.

\bibitem{lattner2021mlir}
Chris Lattner, Mehdi Amini, Uday Bondhugula, Albert Cohen, Andy Davis, Jacques Pienaar, River Riddle, Tatiana Shpeisman, Nicolas Vasilache, and Oleksandr Zinenko.
\newblock Mlir: Scaling compiler infrastructure for domain specific computation.
\newblock In {\em 2021 IEEE/ACM International Symposium on Code Generation and Optimization (CGO)}, pages 2--14. IEEE, 2021.

\bibitem{liu2023llava}
Haotian Liu, Chunyuan Li, Qingyang Wu, and Yong~Jae Lee.
\newblock Visual instruction tuning.
\newblock In {\em NeurIPS}, 2023.

\bibitem{osama2023streamkworkcentricparalleldecomposition}
Muhammad Osama, Duane Merrill, Cris Cecka, Michael Garland, and John~D. Owens.
\newblock Stream-k: Work-centric parallel decomposition for dense matrix-matrix multiplication on the gpu, 2023.

\bibitem{paszke2019pytorch}
Adam Paszke, Sam Gross, Francisco Massa, Adam Lerer, James Bradbury, Gregory Chanan, Trevor Killeen, Zeming Lin, Natalia Gimelshein, Luca Antiga, et~al.
\newblock Pytorch: An imperative style, high-performance deep learning library.
\newblock {\em Advances in neural information processing systems}, 32, 2019.

\bibitem{radford2021learning-clip}
Alec Radford, Jong~Wook Kim, Chris Hallacy, Aditya Ramesh, Gabriel Goh, Sandhini Agarwal, Girish Sastry, Amanda Askell, Pamela Mishkin, Jack Clark, Gretchen Krueger, and Ilya Sutskever.
\newblock Learning transferable visual models from natural language supervision, 2021.

\bibitem{radford2022robust-whisper}
Alec Radford, Jong~Wook Kim, Tao Xu, Greg Brockman, Christine McLeavey, and Ilya Sutskever.
\newblock Robust speech recognition via large-scale weak supervision, 2022.

\bibitem{ragan2013halide}
Jonathan Ragan-Kelley, Connelly Barnes, Andrew Adams, Sylvain Paris, Fr{\'e}do Durand, and Saman Amarasinghe.
\newblock Halide: a language and compiler for optimizing parallelism, locality, and recomputation in image processing pipelines.
\newblock {\em Acm Sigplan Notices}, 48(6):519--530, 2013.

\bibitem{torchfx}
James Reed, Zachary DeVito, Horace He, Ansley Ussery, and Jason Ansel.
\newblock torch.fx: Practical program capture and transformation for deep learning in python.
\newblock In D.~Marculescu, Y.~Chi, and C.~Wu, editors, {\em Proceedings of Machine Learning and Systems}, volume~4, page 638–651, 2022.

\bibitem{Roesch_2018}
Jared Roesch, Steven Lyubomirsky, Logan Weber, Josh Pollock, Marisa Kirisame, Tianqi Chen, and Zachary Tatlock.
\newblock Relay: a new {IR} for machine learning frameworks.
\newblock In {\em Proceedings of the 2nd {ACM} {SIGPLAN} International Workshop on Machine Learning and Programming Languages}. {ACM}, jun 2018.

\bibitem{roesch2020thesis}
Jared~Graham Roesch.
\newblock {\em Principled Optimization Of Dynamic Neural Networks}.
\newblock PhD thesis, University of Washington, 2020.

\bibitem{rozière2024code}
Baptiste Rozière, Jonas Gehring, Fabian Gloeckle, Sten Sootla, Itai Gat, Xiaoqing~Ellen Tan, Yossi Adi, Jingyu Liu, Romain Sauvestre, Tal Remez, Jérémy Rapin, Artyom Kozhevnikov, Ivan Evtimov, Joanna Bitton, Manish Bhatt, Cristian~Canton Ferrer, Aaron Grattafiori, Wenhan Xiong, Alexandre Défossez, Jade Copet, Faisal Azhar, Hugo Touvron, Louis Martin, Nicolas Usunier, Thomas Scialom, and Gabriel Synnaeve.
\newblock Code llama: Open foundation models for code, 2024.

\bibitem{shaikha2017dps}
Amir Shaikhha, Andrew Fitzgibbon, Simon Peyton~Jones, and Dimitrios Vytiniotis.
\newblock Destination-passing style for efficient memory management.
\newblock In {\em Proceedings of the 6th ACM SIGPLAN International Workshop on Functional High-Performance Computing}, FHPC 2017, page 12–23, New York, NY, USA, 2017. Association for Computing Machinery.

\bibitem{shao2022tensor}
Junru Shao, Xiyou Zhou, Siyuan Feng, Bohan Hou, Ruihang Lai, Hongyi Jin, Wuwei Lin, Masahiro Masuda, Cody~Hao Yu, and Tianqi Chen.
\newblock Tensor program optimization with probabilistic programs.
\newblock {\em Advances in Neural Information Processing Systems}, 35:35783--35796, 2022.

\bibitem{MLSYS2021_nimble}
Haichen Shen, Jared Roesch, Zhi Chen, Wei Chen, Yong Wu, Mu~Li, Vin Sharma, Zachary Tatlock, and Yida Wang.
\newblock Nimble: Efficiently compiling dynamic neural networks for model inference.
\newblock In A.~Smola, A.~Dimakis, and I.~Stoica, editors, {\em Proceedings of Machine Learning and Systems}, volume~3, pages 208--222, 2021.

\bibitem{Thakkar_CUTLASS_2023}
Vijay Thakkar, Pradeep Ramani, Cris Cecka, Aniket Shivam, Honghao Lu, Ethan Yan, Jack Kosaian, Mark Hoemmen, Haicheng Wu, Andrew Kerr, Matt Nicely, Duane Merrill, Dustyn Blasig, Fengqi Qiao, Piotr Majcher, Paul Springer, Markus Hohnerbach, Jin Wang, and Manish Gupta.
\newblock {CUTLASS}, jan 2023.

\bibitem{tillet2019triton}
Philippe Tillet, Hsiang-Tsung Kung, and David Cox.
\newblock Triton: an intermediate language and compiler for tiled neural network computations.
\newblock In {\em Proceedings of the 3rd ACM SIGPLAN International Workshop on Machine Learning and Programming Languages}, pages 10--19, 2019.

\bibitem{touvron2023llama}
Hugo Touvron, Louis Martin, Kevin Stone, Peter Albert, Amjad Almahairi, Yasmine Babaei, Nikolay Bashlykov, Soumya Batra, Prajjwal Bhargava, Shruti Bhosale, Dan Bikel, Lukas Blecher, Cristian~Canton Ferrer, Moya Chen, Guillem Cucurull, David Esiobu, Jude Fernandes, Jeremy Fu, Wenyin Fu, Brian Fuller, Cynthia Gao, Vedanuj Goswami, Naman Goyal, Anthony Hartshorn, Saghar Hosseini, Rui Hou, Hakan Inan, Marcin Kardas, Viktor Kerkez, Madian Khabsa, Isabel Kloumann, Artem Korenev, Punit~Singh Koura, Marie-Anne Lachaux, Thibaut Lavril, Jenya Lee, Diana Liskovich, Yinghai Lu, Yuning Mao, Xavier Martinet, Todor Mihaylov, Pushkar Mishra, Igor Molybog, Yixin Nie, Andrew Poulton, Jeremy Reizenstein, Rashi Rungta, Kalyan Saladi, Alan Schelten, Ruan Silva, Eric~Michael Smith, Ranjan Subramanian, Xiaoqing~Ellen Tan, Binh Tang, Ross Taylor, Adina Williams, Jian~Xiang Kuan, Puxin Xu, Zheng Yan, Iliyan Zarov, Yuchen Zhang, Angela Fan, Melanie Kambadur, Sharan Narang, Aurelien Rodriguez, Robert Stojnic, Sergey Edunov, and Thomas
  Scialom.
\newblock Llama 2: Open foundation and fine-tuned chat models, 2023.

\bibitem{NIPS2017_3f5ee243}
Ashish Vaswani, Noam Shazeer, Niki Parmar, Jakob Uszkoreit, Llion Jones, Aidan~N Gomez, \L~ukasz Kaiser, and Illia Polosukhin.
\newblock Attention is all you need.
\newblock In I.~Guyon, U.~Von Luxburg, S.~Bengio, H.~Wallach, R.~Fergus, S.~Vishwanathan, and R.~Garnett, editors, {\em Advances in Neural Information Processing Systems}, volume~30. Curran Associates, Inc., 2017.

\bibitem{hf_transformers}
Thomas Wolf, Lysandre Debut, Victor Sanh, Julien Chaumond, Clement Delangue, Anthony Moi, Perric Cistac, Clara Ma, Yacine Jernite, Julien Plu, Canwen Xu, Teven Le~Scao, Sylvain Gugger, Mariama Drame, Quentin Lhoest, and Alexander~M. Rush.
\newblock {Transformers: State-of-the-Art Natural Language Processing}.
\newblock pages 38--45. Association for Computational Linguistics, October 2020.

\bibitem{xu2023wizardlm}
Can Xu, Qingfeng Sun, Kai Zheng, Xiubo Geng, Pu~Zhao, Jiazhan Feng, Chongyang Tao, and Daxin Jiang.
\newblock Wizardlm: Empowering large language models to follow complex instructions.
\newblock {\em arXiv preprint arXiv:2304.12244}, 2023.

\bibitem{ye2023sparsetir}
Zihao Ye, Ruihang Lai, Junru Shao, Tianqi Chen, and Luis Ceze.
\newblock Sparsetir: Composable abstractions for sparse compilation in deep learning.
\newblock In {\em Proceedings of the 28th ACM International Conference on Architectural Support for Programming Languages and Operating Systems, Volume 3}, pages 660--678, 2023.

\bibitem{MLSYS2022_f89b79c9}
Bojian Zheng, Ziheng Jiang, Cody~Hao Yu, Haichen Shen, Joshua Fromm, Yizhi Liu, Yida Wang, Luis Ceze, Tianqi Chen, and Gennady Pekhimenko.
\newblock Dietcode: Automatic optimization for dynamic tensor programs.
\newblock In D.~Marculescu, Y.~Chi, and C.~Wu, editors, {\em Proceedings of Machine Learning and Systems}, volume~4, pages 848--863, 2022.

\bibitem{ansor}
Lianmin Zheng, Chengfan Jia, Minmin Sun, Zhao Wu, Cody~Hao Yu, Ameer Haj-Ali, Yida Wang, Jun Yang, Danyang Zhuo, Koushik Sen, Joseph~E. Gonzalez, and Ion Stoica.
\newblock Ansor: Generating {High-Performance} tensor programs for deep learning.
\newblock In {\em 14th USENIX Symposium on Operating Systems Design and Implementation (OSDI 20)}, pages 863--879. USENIX Association, November 2020.

\bibitem{zheng2022astitch}
Zhen Zheng, Xuanda Yang, Pengzhan Zhao, Guoping Long, Kai Zhu, Feiwen Zhu, Wenyi Zhao, Xiaoyong Liu, Jun Yang, Jidong Zhai, et~al.
\newblock Astitch: enabling a new multi-dimensional optimization space for memory-intensive ml training and inference on modern simt architectures.
\newblock In {\em Proceedings of the 27th ACM International Conference on Architectural Support for Programming Languages and Operating Systems}, pages 359--373, 2022.

\bibitem{zhu2021disc}
Kai Zhu, Wenyi Zhao, Zhen Zheng, Tianyou Guo, Pengzhan Zhao, Feiwen Zhu, Junjie Bai, Jun Yang, Xiaoyong Liu, Lansong Diao, and Wei Lin.
\newblock Disc: A dynamic shape compiler for machine learning workloads, 2021.

\end{thebibliography}
